\documentclass[10pt,twocolumn,letterpaper]{article}

\usepackage{cvpr}
\usepackage{times}
\usepackage{epsfig}
\usepackage{graphicx}
\usepackage{amsmath}
\usepackage{amssymb}
\usepackage{multirow}
\usepackage{amssymb}
\usepackage{mathrsfs}
\usepackage{url}
\usepackage{dsfont}

\usepackage[pagebackref=true,breaklinks=true,letterpaper=true,colorlinks,bookmarks=false]{hyperref}

\cvprfinalcopy 

\def\cvprPaperID{4839} 

\begin{document}

\title{Deep Image Deraining Via Intrinsic Rainy Image Priors and \\ Multi-scale Auxiliary Decoding}

\author{Yinglong Wang{$^{\dag}$}, Chao Ma{$\ddag$} Bing Zeng{$^{\dag}$}\\
$^{\dag}$University of Electronic Science and Technology of China\\
$^{\ddag}$The Shanghai Jiaotong University\\}

\maketitle

\begin{abstract}
Different rain models and novel network structures have been proposed to remove rain streaks from single rainy images. In this work, we bring attention to the intrinsic priors and multi-scale features of the rainy images, and develop several intrinsic loss functions to train a CNN deraining network. We first study the sparse priors of rainy images, which have been verified to preserve unbroken edges in image decomposition. However, its mathematical formulation usually leads to an intractable solution, we propose quasi-sparsity priors to decrease complexity, so that our network can be trained under the supervision of sparse properties of rainy images. Quasi-sparsity supervises network training in different gradient domain which is still ill-posed to decompose a rainy image into rain layer and background layer. We develop another $L_1$ loss based on the intrinsic low-value property of rain layer to restore image contents together with the commonly-used $L_1$ similarity loss. Multi-scale features are further explored via a multi-scale auxiliary decoding structure to show which kinds of features contribute the most to the deraining task, and the corresponding multi-scale auxiliary loss improves the deraining performance further. In our network, more efficient group convolution and feature sharing are utilized to obtain an one order of magnitude improvement in network running speed. The proposed deraining method performs favorably against state-of-the-art deraining approaches.
\end{abstract}

\section{Introduction}

\begin{figure}[t]
\centering
\begin{minipage}{0.32\linewidth}
\centering{\includegraphics[width=1\linewidth]{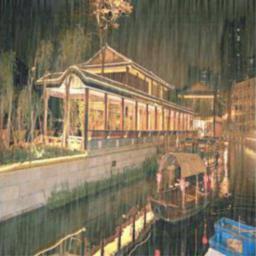}}
\centerline{(a)}
\end{minipage}
\hfill
\begin{minipage}{.32\linewidth}
\centering{\includegraphics[width=1\linewidth]{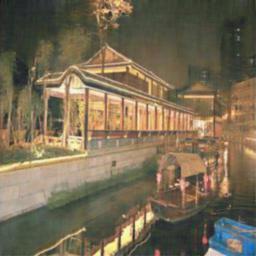}}
\centerline{(b)}
\end{minipage}
\hfill
\begin{minipage}{.32\linewidth}
\centering{\includegraphics[width=1\linewidth]{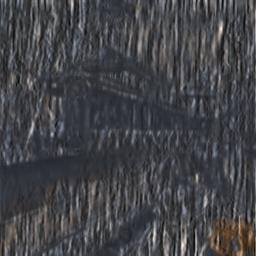}}
\centerline{(c)}
\end{minipage}
\vfill
\begin{minipage}{0.32\linewidth}
\centering{\includegraphics[width=1\linewidth]{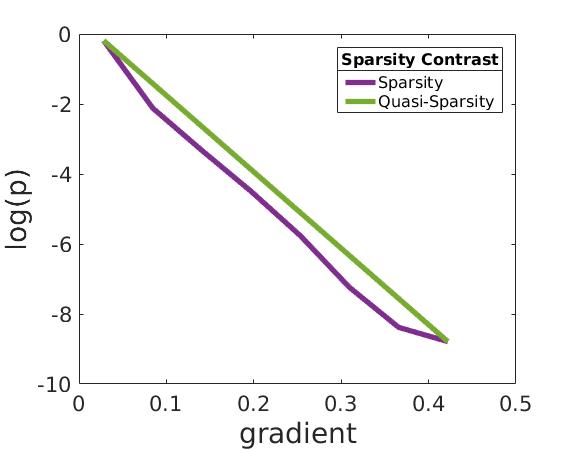}}
\centerline{(d)}
\end{minipage}
\hfill
\begin{minipage}{.32\linewidth}
\centering{\includegraphics[width=1\linewidth]{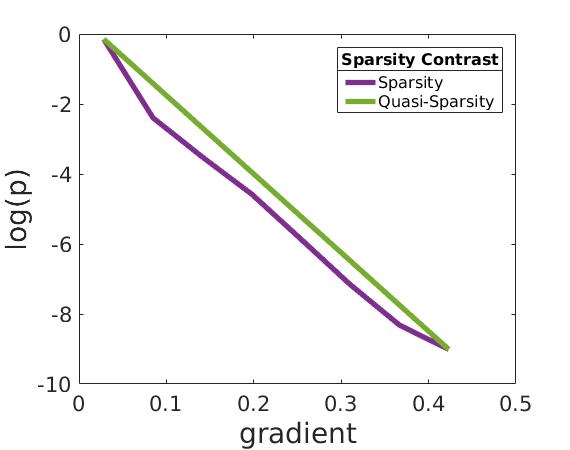}}
\centerline{(e)}
\end{minipage}
\hfill
\begin{minipage}{.32\linewidth}
\centering{\includegraphics[width=1\linewidth]{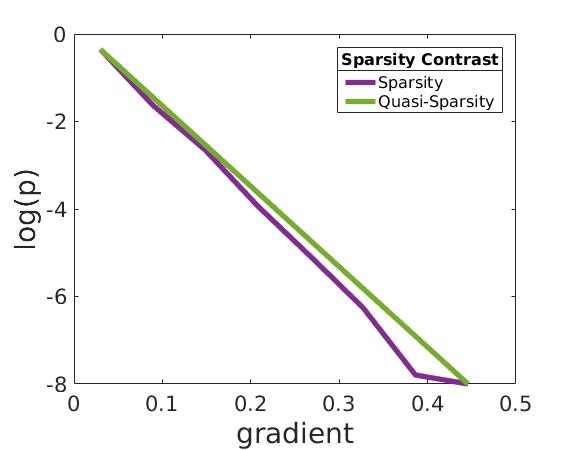}}
\centerline{(f)}
\end{minipage}
\caption{(a) Input. (b) Deraining result. (c) Removed rain streak layer. (d)-(f) Sparse distributions of input, deraining result and rain streak layer. These three variants obey sparse distribution.}
\label{fig:teaser}
\end{figure}

Rain usually impairs images through two ways: 1) large raindrops are always imaged as apparent rain streaks, which distort image textures and occlude the background behind them; 2) dense tiny raindrops accumulate together and generate a layer of haze-like effect which lowers the image contrast and merge some fine image details \cite{Wang_yl_2019_arxiv,Yang_2017_CVPR,Li_rt_2019_CVPR}. Considering that removing rain streaks is still an unsolved issue in some realistic scenarios, special attentions are still needed to push the further solution of this problem. In this work, we focus on removing rain streaks from the perspective of the intrinsic priors of rainy images.

Conventional methods usually decompose rain streaks from rainy image via learning an over-complete dictionary \cite{Wang_2017_TIP,Chen_2014_CSVT,Kang_2012_TIP}. Learning a dictionary is time consuming \cite{Mairal_2010_JMLR}, and the deraining performance is limited due to the heuristic appearance characteristics of rain streaks. Recently, via exploring high-level features of rainy images, deep learning has achieved state-of-the-art deraining performance in not only the image recovery but also the running speed \cite{Fu_2017_CVPR,Fu_2017_TIP,Yang_2017_CVPR,Zhang_2018_CVPR,Li_2018_ECCV}. Majority of deep learning based deraining methods model the observed rainy image $\mathbf{I}$ as a summation of a background layer and rainy layers.
Some works build more complete models to formulate $\mathbf{I}$, but they are eventually simplified into residual model by a CNN network \cite{Yang_2017_CVPR}. A difference exists in \cite{Li_rt_2019_CVPR}, in which atmospheric light and transmission are introduce and learned to model rainy image more completely.

Except for the rain models, existing methods mainly focus on designing novel network structures to capture more effective deep features of rain streaks to promote performance, for example, density guided DID-MDN \cite{Zhang_2018_CVPR}, location guided JORDER \cite{Yang_2017_CVPR,Yang_wh_2019_TPAMI}, residual based DDN \cite{Fu_2017_CVPR}, depth guided DAF-Net \cite{Hu_xw_2019_CVPR}, recurrent RESCAN \cite{Li_2018_ECCV}, progressive PReNet \cite{Ren_dw_2019_CVPR}, \etc. While the training of networks basically relies on minimizing the deviation between the deraining result and ground truth background, including the $L_2$ loss MSE \cite{Zhang_2018_CVPR,Fu_2017_CVPR,Yang_2017_CVPR}, and $L_1$ loss MAE \cite{Li_2018_MM} to restore image contents and SSIM loss \cite{Wang_2004_TIP} to repair image textures. $L_2$ loss converges at the arithmetic mean of the observations, which is a global statistical value. Hence, $L_2$ correlates poorly with image quality which is sensitive to the local characteristics of images \cite{Lehtinen_2018_ICML}. The local metric $L_1$ reaches its minimum at the median of the observations, which is better for restoring local characteristics \cite{Zhao_2017_TCI}. $L_1$ loss may generate outlier content during the image restoration \cite{Lehtinen_2018_ICML}. Both $L_1$ and $L_2$ are utilized to measure the similarity of deraining result and the ground truth. Few methods take the intrinsic properties of rainy image into consideration to restore the clear background from the degraded version to alleviate the limitations of these standard losses.

We rely on the intrinsic priors of rainy images to derive new cost functions to train a CNN deraining network. We first explore the sparse properties of rainy images by a robust statistical experiment. Sparsity priors of natural images has been verified to boost the unbroken factorization of image contents \cite{Levin_2007_PAMI}. This property is useful to decompose intact rain streaks into the rain layer and keep the image details in background layer. We develop a new \textit{quasi-sparsity priors} by reasonably relaxing the sparsity degree to formulate the sparsity of rainy images. Quasi-sparsity possesses simpler form, which eases the training of a CNN network via the maximum likelihood estimation (MLE) based on quasi-sparse distribution. Deraining issue is still ill-posed only via quasi-sparse constraints, which just separates the textures of rain streaks and background in sparse domain. Hence, a content loss measuring the deviation in spatial domain is added to separate the contents of rain and background complementarily. The low-value prior of rain layer, \ie, the non-rain areas in rain layer have values that closes to $0$, is formulated as a $L_1$ detail loss to preserve fine details of background further. Fig. \ref{fig:teaser} is an example of decomposing a real-world rainy image into rain layer and background layer and their corresponding sparsity curves.

Second, rainy features with different scales has been used in previous deraining works \cite{Zhang_2018_CVPR,Yang_2017_CVPR}, but few methods bring special attention to studying what kinds of feature is most favorable to deraining task. In our work, We also extract the multi-scale features of rain streaks due to their various shapes and sizes. Accordingly, a novel multi-scale auxiliary decoder structure is introduced in our network to boost network to generate more effective deraining features via multi-scale auxiliary cost function. Moreover, we study the performance of deraining results decoded from different auxiliary decoder in detail to show which kinds of features contribute most to deraining.

\begin{figure*}
\centering
\begin{minipage}{1\linewidth}
\centering{\includegraphics[width=1\linewidth]{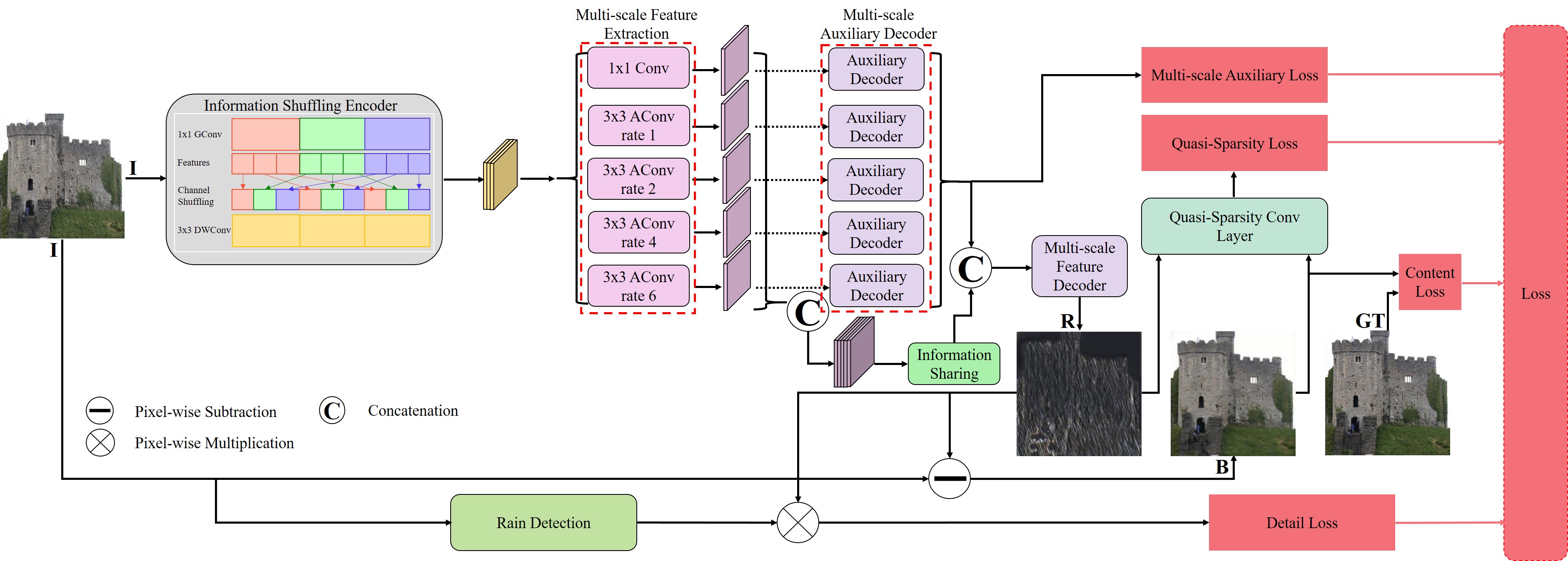}}
\centerline{}
\end{minipage}
\caption{Pipeline of our method. A stack of $12$ ShuffleNet units constitute the backbone of our network. Multi-scale Auxiliary Decoder is introduced to separately decode deraining results from different scale features. We focus on the training of our network, the intrinsic quasi-sparsity loss, detail loss based on the low-value property of rainy layer and another two similarity metrics (content loss and multi-scale auxiliary loss) work together to train our network.}
\label{fig:pipeline}
\end{figure*}

Third, our network contains many convolution layer, we decrease the parameters of our network by applying group convolution instead, which is also more efficient than the common convolution to obtain a fast processing speed. Inspired by the shuffling operation in \cite{Zhang_2018_CVPR_Shuffle}, we exchange the features from different groups in the auxiliary and the main decoder, and also the features from the different auxiliary decoder to fuse different rainy features. Ablation studies illustrate that feature exchanging further enhance deraining performance and running speed. The outline of our network structure is in Fig. \ref{fig:pipeline}.

We summarize our main contributions in the following.
\begin{itemize}
\item 
We study the sparsity of rainy images via robust statistical experiment to show the influence of rain on natural images in sparse domain, which provides theoretical foundation for the network training by using maximum likelihood estimation based on sparse distribution. 
\item
We develop quasi-sparse priors to formulate the sparsity of rainy images. The form of quasi-sparsity simplify the network training based on sparse distribution and a tractable $L_{1}$ cost function is derived to restore background texture in sparse domain. Besides a similarity metric in spatial domain, the low-value priors of rainy layer is formulated as $L_{1}$ minimization cost function to restore image details.
\item
We introduce a novel multi-scale auxiliary decoding CNN network based on the efficient group convolution and feature exchanging/sharing of different groups and scales. We study the influence of different kinds of features on deraining performance and determine the feature scale which contributes most to deraining. Corresponding to auxiliary decoder, a multi-scale auxiliary cost function is derived to help network extract more favorable deraining features.
\item 
Compared to the state-of-the-art deraining works, our method obtains better deraining performance. Our running speed is improved greatly compared to the existing deep learning based works ( $10^{-3}$ order of magnitudes).
\end{itemize}

\section{Related Works}
\label{sec:RelatedWorks}

Dictionary learning \cite{Mairal_2010_JMLR} is first used to remove rain streaks from single image by decomposing image content into multiple different layers \cite{Kang_2012_TIP,Chen_2014_CSVT,Wang_2017_TIP,Wang_2016_ICIP}. Recently, deep learning improves the deraining performance substantially. Based on the rain model, prevalent deep learning deraining methods can be categorized into three aspect: direct learning method, residual model method and scattering model methods. Direct learning methods learn rain-free background directly from the observed rainy images. In \cite{Wang_ty_2019_CVPR} a dataset is first built via combining temporal priors and human supervision, based on which a SPANet is trained to solve the random rain streaks in a local-to-global way.

Residual methods decompose rainy image into a rainy layer and rain-free background layer. In \cite{Fu_2017_TIP}, a DerainNet is trained in high-frequency domain to restore image details, so that interference from background can be reduced. A deep detail network based on ResNet \cite{He_2015_CVPR} is also trained in high-pass to reduce the mapping range from input to output, so that the learning process becomes easier \cite{Fu_2017_CVPR}. In \cite{Yang_2017_CVPR,Yang_wh_2019_TPAMI}, a new rain model is introduced to model the apparent rain streaks and the veiling effect caused by the accumulation of rain streaks. But the atmospheric light and transmission of veiling effect are not explicitly predicted and the rainy image is finally decomposed into a rain layer and background layer by their JORDER network. Moreover, a binary map is learnt to locate rain streaks to guide the JORDER. In \cite{Zhang_2018_CVPR}, the density of rain streaks is evaluated with which a multi-stream densely connected DID-MDN structure is trained to better characterize rain streaks with various shape and size. Li \etal decompose rain streaks in single images into several rain layers, then a recurrent neural network RESCAN is trained to remove rain streaks state-wisely \cite{Li_2018_ECCV}. Hu \etal study the relationship between visual effect of rain and scene depth, based on which fog that contains depth information is introduced to model the formation of rainy images and to guide the training of their end-to-end network \cite{Hu_xw_2019_CVPR}. In \cite{Ren_dw_2019_CVPR}, Ren \etal rethink the network structure, input and output of network, and the loss functions and a better and simpler deraining baseline is proposed.

In the scattering model methods, Li \etal render the ground truth for atmospheric light, rain streaks and transmission of vapor to remove rain streaks as well as vapor effect. Different from existing approaches, we bring attention to the intrinsic priors of rainy images and focus on the deraining  cost function.

\begin{figure}[t]
\begin{minipage}{0.43\linewidth}
\centering{\includegraphics[width=1\linewidth]{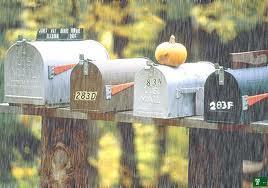}}
\centerline{(a)}
\end{minipage}
\hfill
\begin{minipage}{.43\linewidth}
\centering{\includegraphics[width=1\linewidth]{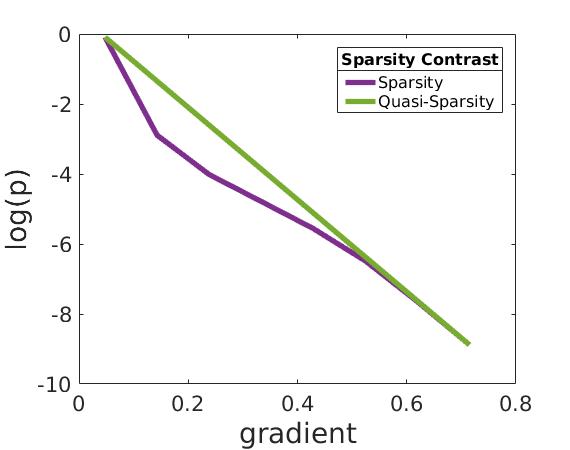}}
\centerline{(b)}
\end{minipage}
\vfill
\begin{minipage}{0.43\linewidth}
\centering{\includegraphics[width=1\linewidth]{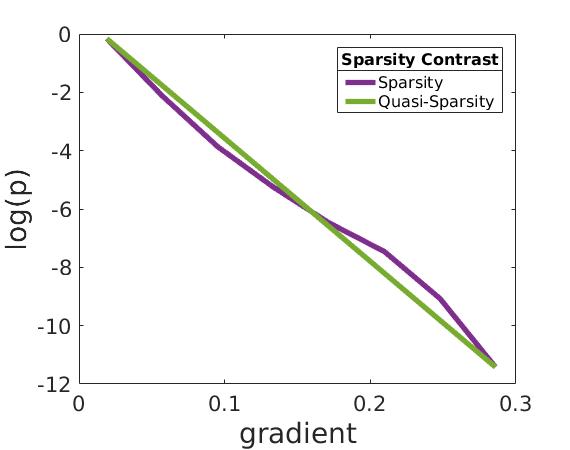}}
\centerline{(c)}
\end{minipage}
\hfill
\begin{minipage}{.43\linewidth}
\centering{\includegraphics[width=1\linewidth]{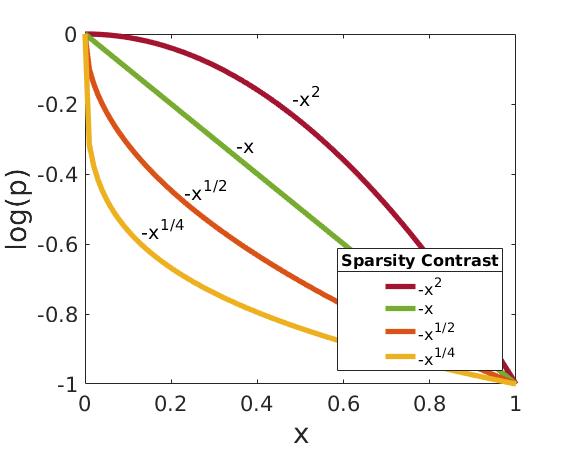}}
\centerline{(d)}
\end{minipage}
\caption{(a) A rainy image example. (b) Log-histogram after (a) is filtered by gradient mask. (c) Log-histogram of small percentage of rainy images which are non-sparse. (d) Log-probability of several common distributions.}
\label{fig:sparsity_verify}
\end{figure}

\begin{figure}
\centering
\begin{minipage}{1\linewidth}
\centering{\includegraphics[width=1\linewidth]{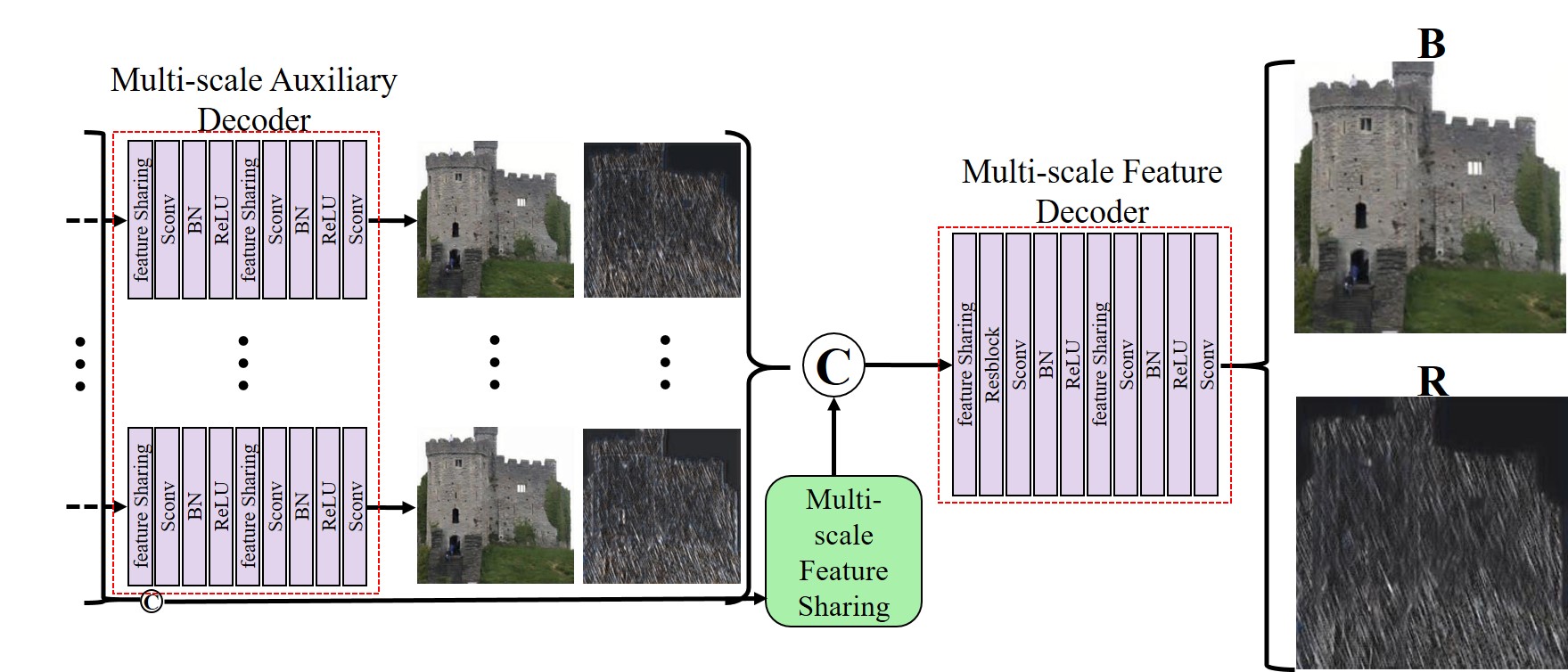}}
\centerline{}
\end{minipage}
\caption{This figure shows the details of our multi-scale auxiliary decoder and multi-scale feature decoder. We let different convolution groups exchange information in single auxiliary decoder. The features from different scales are also exchanged to fuse different rainy features.}
\label{fig:decoder}
\end{figure}

\section{Quasi-sparsity priors}
\label{sec:sparsity}

We first study the sparsity of rainy image by implementing an extensive statistics. Sparsity is favorable to the intact factorization of image content \cite{Levin_2007_PAMI}. But the formulation of sparsity priors are complex \cite{Levin_2007_PAMI,Wainwright_2000_ICIP}, which hinders their utility in more extensive computer vision tasks \ie, deraining. We formulate sparsity priors of rainy images as quasi-sparse priors to obtain a tractable solution.

\subsection{Study the Sparsity of Rainy Images}

In \cite{Levin_2007_PAMI}, Levin and Weiss shows a robust fact that the log-histograms of filtered natural images are below the straight line connecting the minimal and maximal values. This intrinsic property of natural images is named as sparsity priors. In this work, we study the rainy images to verify their sparsity in a statistical way. We collect $1000$ real-world images and $9000$ synthetic images and calculate their log-histogram. By statistics, $93.7 \%$ of them keep sparsity priors, an example is shown in Fig. \ref{fig:sparsity_verify}(a)(b). The log-histogram of the remaining part is displayed in Fig. \ref{fig:sparsity_verify}(c). By experiment, we find this minor part of non-sparse distribution mainly caused by the severe low contrast of these rainy images. Irrespective of such small part, rainy images are sparse from a statistical perspective.

\subsection{Quasi-Sparse Formulation of Rainy Images}
\label{sec:quasi_sparsity_priors}

The tractable Gaussian and Laplacian distributions are non-sparse as shown in Fig. \ref{fig:sparsity_verify}(d). Single Gaussian is above the straight line. Infinite Gaussian distributions with different parameters are added together to model the sparsity of images in \cite{Wainwright_2000_ICIP}, which simultaneously brings complex solution. Laplacian right results in the straight line.
Levin and Weiss \cite{Levin_2007_PAMI} fit the sparsity priors by adding two different  Laplacian distributions:
\begin{equation} \label{eq:original_sparsity}
P(x) = \frac{\pi_{1}}{2s_{1}}e^{- \vert x \vert/s_{1}} + \frac{\pi_{2}}{2s_{2}}e^{- \vert x \vert/s_{2}},
\end{equation}
where $\pi_1$, $\pi_2$, $s_1$ and $s_2$ are parameters in Laplacian distributions. The complexity is substantial decreased, but still intractable to optimize a CNN network via maximum likelihood estimation based on sparsity priors.

We further relax the degree of sparsity by setting $s_{1}=s_{2}=s$ and $\pi_{1}=\pi_{2}=\pi$ in Eq. \eqref{eq:original_sparsity}:
\begin{equation} \label{eq:simplify_sparsity}
P_{q}(x) = \frac{\pi}{s}e^{- \vert x \vert/s},
\end{equation}
\ie, the sparsity of rainy images is formulated as single Laplacian distribution, which is the borderline of sparsity and non-sparsity. We call single Laplacian as \textit{quasi-sparsity priors}. There is only one exponent in quasi-sparsity priors, a solvable $L_1$ cost can be easily derived by logarithmic MLE, so that CNN network can be trained based on the intrinsic image sparsity priors. Moreover, quasi-sparsity priors are closer to the non-sparse distribution of minority of rainy images in Fig. \ref{fig:sparsity_verify}(c) which facilitates the handling of these rainy images together with sparse rainy images, even a better solution. The quasi-sparse distribution of rain image $\mathbf{I}$ can be written as:
\begin{equation}\label{eq:laplacian_approximation}
P_{q}(\mathbf{I})=\prod_{i, k}P_{q}(\omega_{i, k} \ast \mathbf{I} )
\end{equation}
where $\omega_{i,k}$ is the $k^{th}$ filter which centered
at $i^{th}$ pixel. $\ast$ is convolution. The filters
with two orientations (horizontal and vertical) and two degrees (the
first derivative and the second derivative) are used here to constitute the quasi-sparsity priors of rainy images.

\section{Proposed Method}
\label{sec:our_method}

We train a CNN network to predict rain streak layer $\mathbf{R}$ from rainy image $\mathbf{I}$, rain free background $\mathbf{B}$ is obtained by subtracting $\mathbf{R}$ from $\mathbf{I}$:
\begin{equation}\label{eq:network}
\mathbf{R} =\mathcal{S}(\mathbf{I}),
\end{equation}
\begin{equation}\label{eq:background_layer}
\mathbf{B} =\mathbf{I}-\mathbf{R},
\end{equation}
where $\mathcal{S}(\cdot)$ denotes our network inference. Because our network is trained based on quasi-sparsity priors, we call our network Quasi-Sparse deraining Network (QSNet). We first show the network of our QSNet, then derive intrinsic cost functions to train our network.

\begin{figure}[t]
\centering
\begin{minipage}{0.32\linewidth}
\centering{\includegraphics[width=1\linewidth]{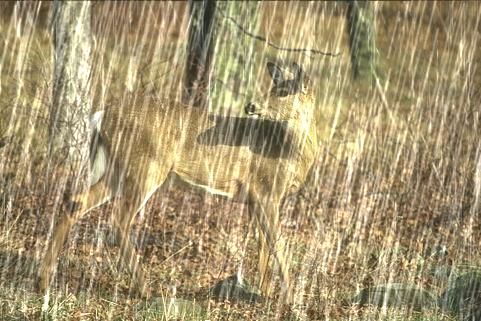}}
\centerline{(a)}
\end{minipage}
\hfill
\begin{minipage}{.32\linewidth}
\centering{\includegraphics[width=1\linewidth]{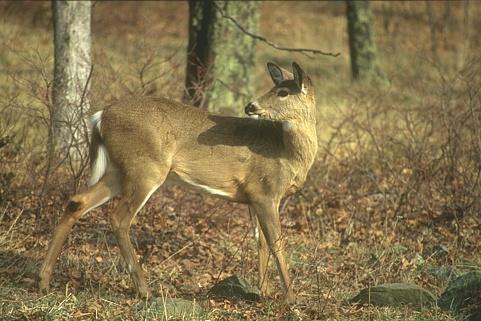}}
\centerline{(b)}
\end{minipage}
\hfill
\begin{minipage}{.32\linewidth}
\centering{\includegraphics[width=1\linewidth]{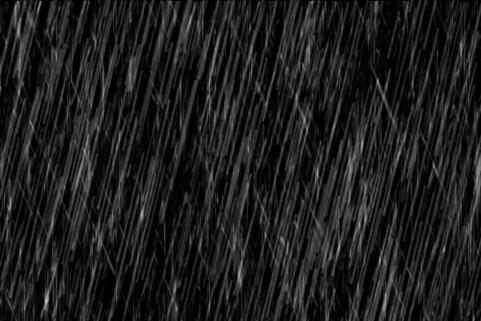}}
\centerline{(c)}
\end{minipage}
\vfill
\begin{minipage}{0.32\linewidth}
\centering{\includegraphics[width=1\linewidth]{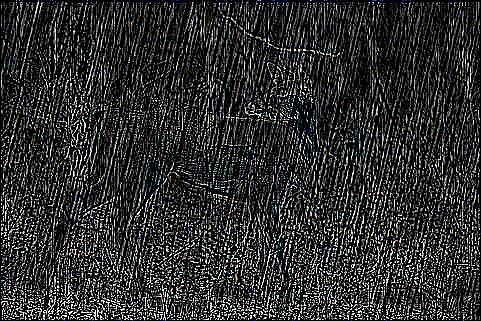}}
\centerline{(d)}
\end{minipage}
\hfill
\begin{minipage}{.32\linewidth}
\centering{\includegraphics[width=1\linewidth]{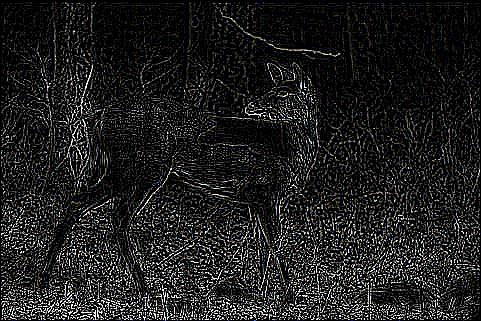}}
\centerline{(e)}
\end{minipage}
\hfill
\begin{minipage}{.32\linewidth}
\centering{\includegraphics[width=1\linewidth]{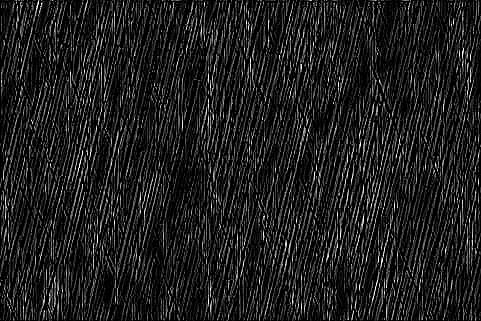}}
\centerline{(f)}
\end{minipage}
\caption{(a) Input rainy image. (b) Background image. (c) Rain streaks. (d)-(f) are first order horizontal gradient of (a)-(c).}
\label{fig:sparse_space_image}
\end{figure}

\subsection{Network Structure}
We show an overview of the pipeline in Fig. \ref{fig:pipeline}. The backbone of our network is a stack of $12$ ShuffleNet units to extract high-level features of rainy images, which is called Information Shuffling Encoder. ShuffleNet units have been verified to have a fast feature extraction speed \cite{Zhang_2018_CVPR_Shuffle}. The network parameters also decrease apparently due to the lightweight group convolution and deep separable convolution. As \cite{Yang_2017_CVPR}, we also use atrous convolution \cite{Chen_2017_PAMI} to generate multi-scale features of rain streaks. Atrous convolution extracts multi-scale features by simply setting proper hyper parameters and obtained features keep the same size as original rainy image, which avoids the down sampling and up sampling using other multi-scale structures, \eg, spatial pyramid pooling \cite{He_2015_arxiv}. A pointwise convolution and $4$ atrous convolutions with different atrous rates constitute our Multi-scale Feature Extraction structure as shown in Fig. \ref{fig:pipeline}. Large-scale features is not favorable for low-level deraining task sensitive to local details, we select small values $1$, $2$, $4$, $6$ as our atrous rate. Moreover, pointwise convolution acts as a shortcut path to preserve previous single-scale feature.

Our decoder contains Multi-scale Auxiliary Decoder and Multi-scale Feature Decoder (main decoder). Each auxiliary decoder decodes out a rain free image from corresponding scale feature. The decoded rain free images are concatenated together with previous multi-scale features as the input of our main decoder as shown in Fig. \ref{fig:pipeline}, and more details are in Fig. \ref{fig:decoder}. Inspired by the shuffling in ShuffleNet unit, we let different convolution groups in each auxiliary decoder share features mutually to fuse features with equal scale. Features from different scale paths of Multi-scale Feature Extraction unit also exchange information before being input into the main decoder.

\begin{figure*}[!t]
\begin{center}
\begin{minipage}{0.105\linewidth}
\centering{\includegraphics[width=1\linewidth]{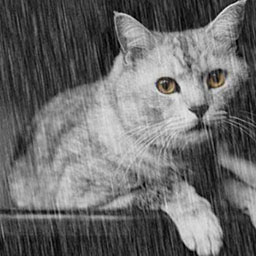}}
\end{minipage}
\hfill
\begin{minipage}{0.105\linewidth}
\centering{\includegraphics[width=1\linewidth]{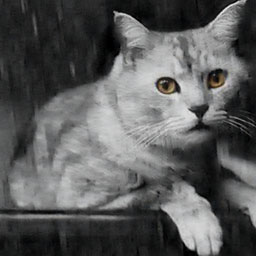}}
\end{minipage}
\hfill
\begin{minipage}{0.105\linewidth}
\centering{\includegraphics[width=1\linewidth]{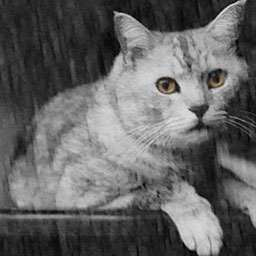}}
\end{minipage}
\hfill
\begin{minipage}{0.105\linewidth}
\centering{\includegraphics[width=1\linewidth]{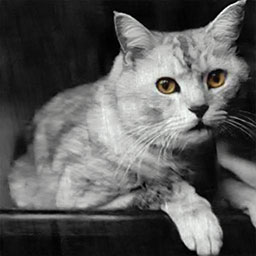}}
\end{minipage}
\hfill
\begin{minipage}{0.105\linewidth}
\centering{\includegraphics[width=1\linewidth]{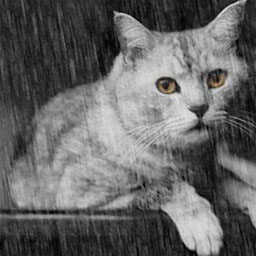}}
\end{minipage}
\hfill
\begin{minipage}{0.105\linewidth}
\centering{\includegraphics[width=1\linewidth]{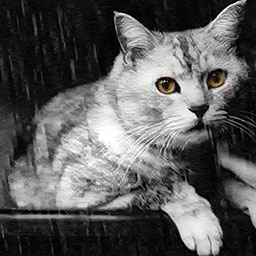}}
\end{minipage}
\hfill
\begin{minipage}{0.105\linewidth}
\centering{\includegraphics[width=1\linewidth]{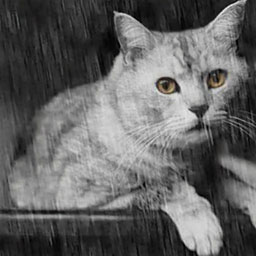}}
\end{minipage}
\hfill
\begin{minipage}{0.105\linewidth}
\centering{\includegraphics[width=1\linewidth]{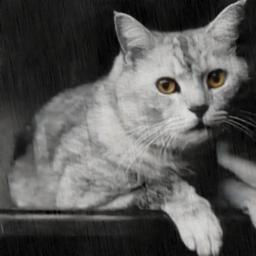}}
\end{minipage}
\hfill
\begin{minipage}{0.105\linewidth}
\centering{\includegraphics[width=1\linewidth]{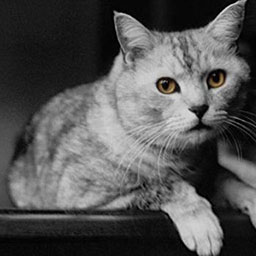}}
\end{minipage}
\vfill
\begin{minipage}{0.105\linewidth}
\centering{\includegraphics[width=1\linewidth]{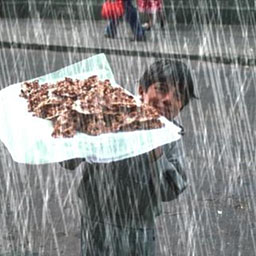}}
\centerline{(a)}
\end{minipage}
\hfill
\begin{minipage}{0.105\linewidth}
\centering{\includegraphics[width=1\linewidth]{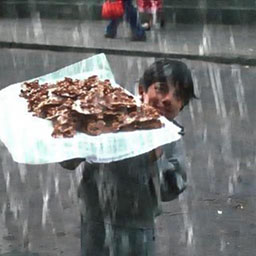}}
\centerline{(b)}
\end{minipage}
\hfill
\begin{minipage}{0.105\linewidth}
\centering{\includegraphics[width=1\linewidth]{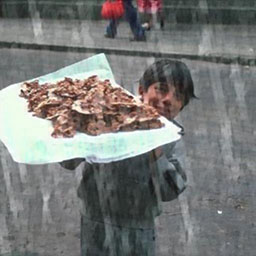}}
\centerline{(c)}
\end{minipage}
\hfill
\begin{minipage}{0.105\linewidth}
\centering{\includegraphics[width=1\linewidth]{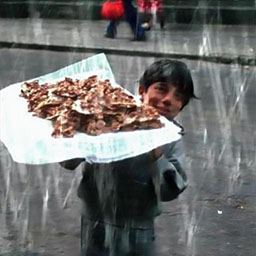}}
\centerline{(d)}
\end{minipage}
\hfill
\begin{minipage}{0.105\linewidth}
\centering{\includegraphics[width=1\linewidth]{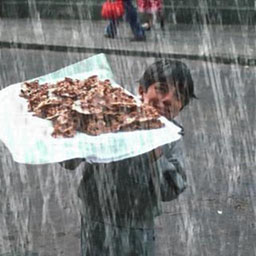}}
\centerline{(e)}
\end{minipage}
\hfill
\begin{minipage}{0.105\linewidth}
\centering{\includegraphics[width=1\linewidth]{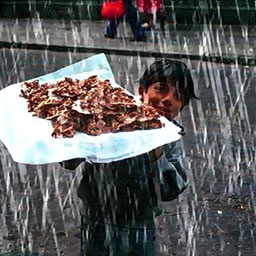}}
\centerline{(f)}
\end{minipage}
\hfill
\begin{minipage}{0.105\linewidth}
\centering{\includegraphics[width=1\linewidth]{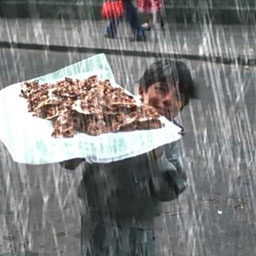}}
\centerline{(g)}
\end{minipage}
\hfill
\begin{minipage}{0.105\linewidth}
\centering{\includegraphics[width=1\linewidth]{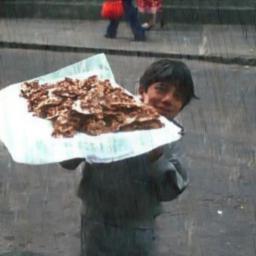}}
\centerline{(h)}
\end{minipage}
\hfill
\begin{minipage}{0.105\linewidth}
\centering{\includegraphics[width=1\linewidth]{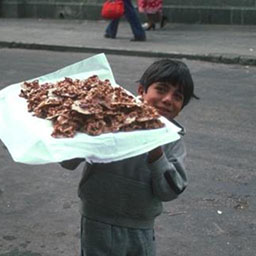}}
\centerline{(i)}
\end{minipage}
\end{center}
\caption{Qualitative comparisons of selected methods and our method on synthetic rainy images. (a) Input. (b)-(h) Deraining results of \cite{Fu_2017_CVPR}, \cite{Li_2018_ECCV}, \cite{Zhang_2018_CVPR}, \cite{Yang_wh_2019_TPAMI}, \cite{Li_rt_2019_CVPR}, \cite{Wang_ty_2019_CVPR} and our method. (i) Ground Truth.}
\label{fig:synthetic_compare}
\end{figure*}

\begin{table*}[]
\centering
\caption{PSNR/SSIM comparisons of selected state-of-the-art and our methods on our three testing datasets.}
\label{tab:psnr_ssim_comparison}
\begin{tabular}{c|cccccc|c}
\hline
\hline
 Methods & \cite{Fu_2017_CVPR} & \cite{Li_2018_ECCV} & \cite{Zhang_2018_CVPR} & \cite{Yang_wh_2019_TPAMI} & \cite{Li_rt_2019_CVPR} & \cite{Wang_ty_2019_CVPR} & Ours \\ \hline
 \hline
 Test-I & 29.14/0.869 & 27.21/0.835 & 25.98/0.869 & 27.44/0.885 & 17.93/0.677 & 28.47/0.858 & \textbf{33.15/0.923} \\ 
 Test-II & 22.17/0.732 & 24.29/0.821 & 20.13/0.716 & 20.46/0.678 & 16.96/0.464 & 18.60/0.623 & \textbf{25.66/0.830} \\
 Test-III & 29.98/0.897 & 26.77/0.832 & 25.53/0.877 & 29.87/0.890 & 18.04/0.610 & 30.55/0.908 & \textbf{33.94/0.938} \\ \hline
 \hline
\end{tabular}
\end{table*}

\begin{figure*}[!t]
\begin{center}
\begin{minipage}{0.118\linewidth}
\centering{\includegraphics[width=1\linewidth]{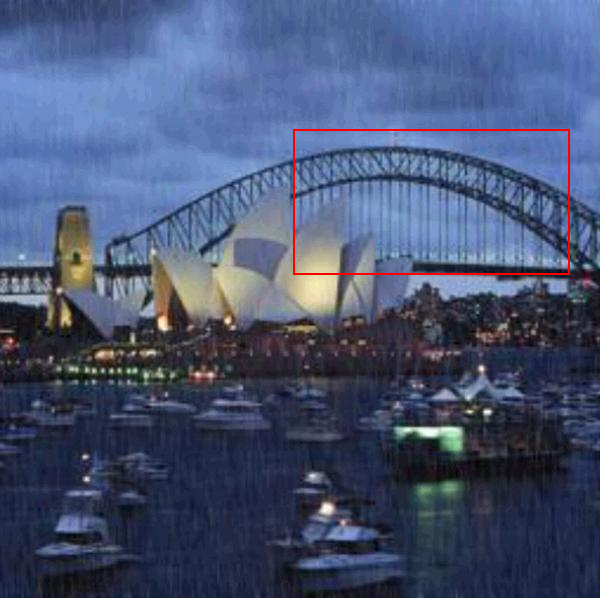}}
\end{minipage}
\hfill
\begin{minipage}{0.118\linewidth}
\centering{\includegraphics[width=1\linewidth]{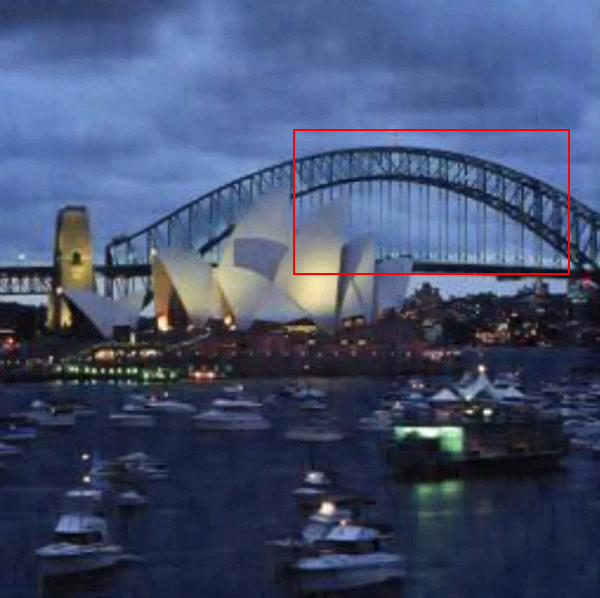}}
\end{minipage}
\hfill
\begin{minipage}{0.118\linewidth}
\centering{\includegraphics[width=1\linewidth]{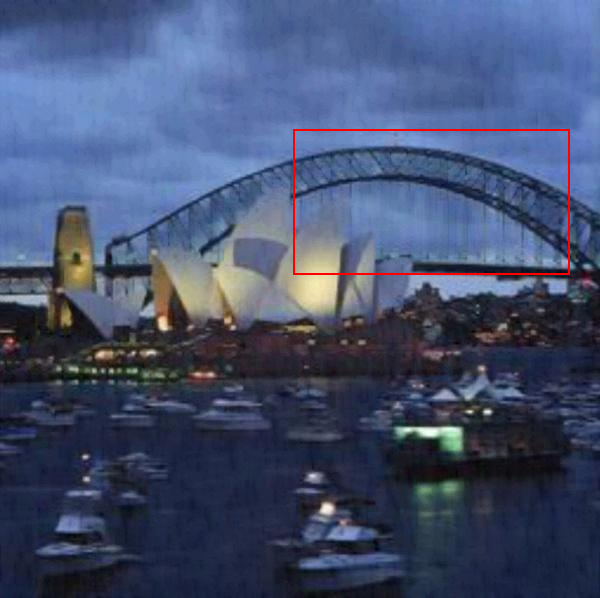}}
\end{minipage}
\hfill
\begin{minipage}{0.118\linewidth}
\centering{\includegraphics[width=1\linewidth]{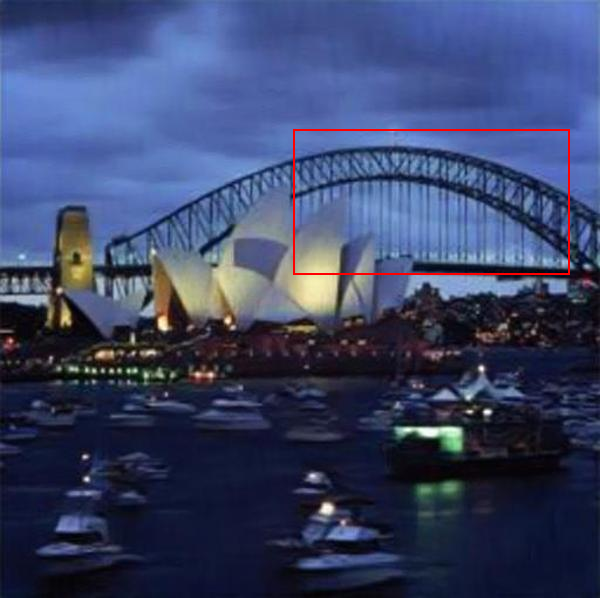}}
\end{minipage}
\hfill
\begin{minipage}{0.118\linewidth}
\centering{\includegraphics[width=1\linewidth]{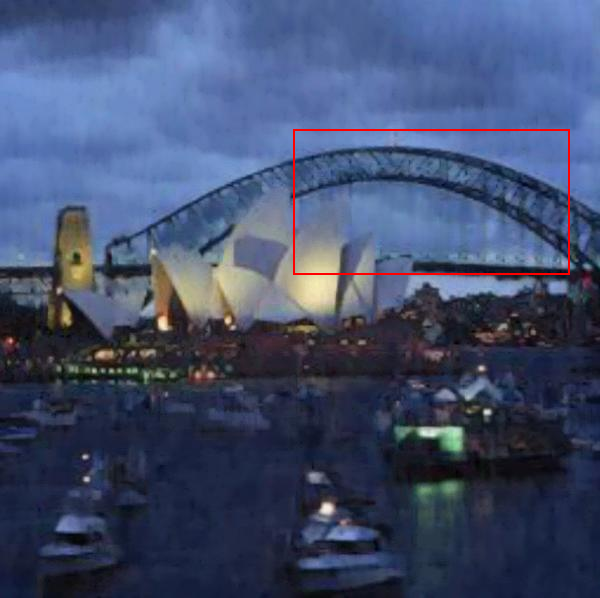}}
\end{minipage}
\hfill
\begin{minipage}{0.118\linewidth}
\centering{\includegraphics[width=1\linewidth]{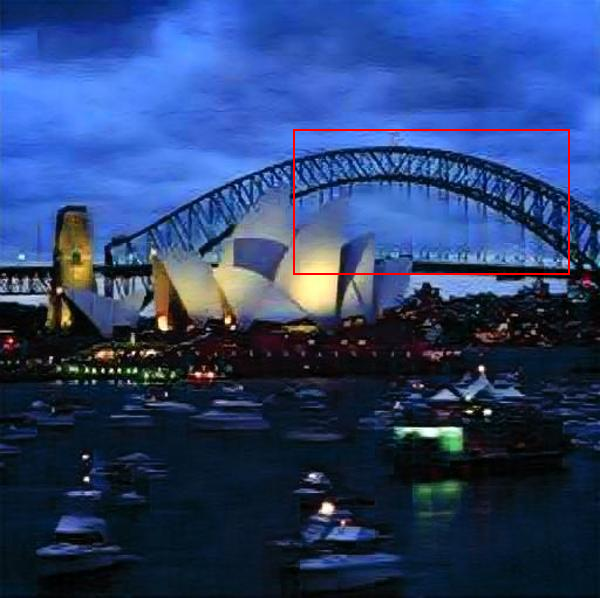}}
\end{minipage}
\hfill
\begin{minipage}{0.118\linewidth}
\centering{\includegraphics[width=1\linewidth]{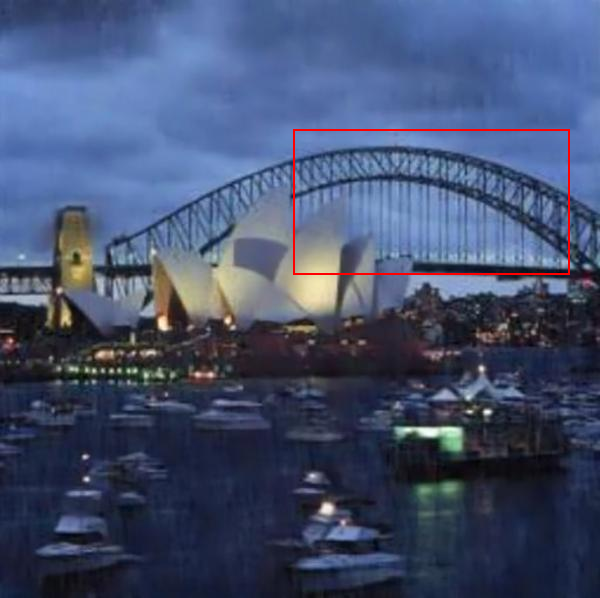}}
\end{minipage}
\hfill
\begin{minipage}{0.118\linewidth}
\centering{\includegraphics[width=1\linewidth]{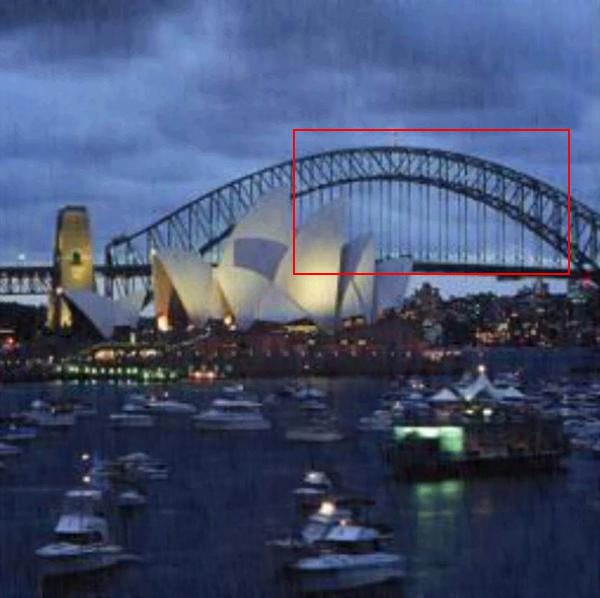}}
\end{minipage}
\vfill
\begin{minipage}{0.118\linewidth}
\centering{\includegraphics[width=1\linewidth]{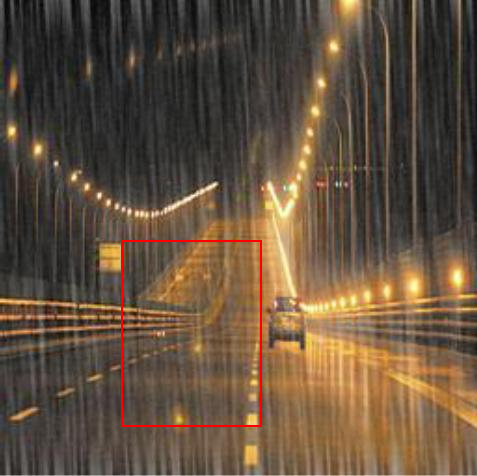}}
\end{minipage}
\hfill
\begin{minipage}{0.118\linewidth}
\centering{\includegraphics[width=1\linewidth]{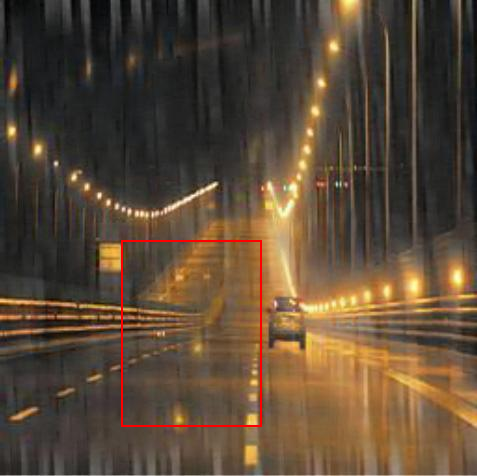}}
\end{minipage}
\hfill
\begin{minipage}{0.118\linewidth}
\centering{\includegraphics[width=1\linewidth]{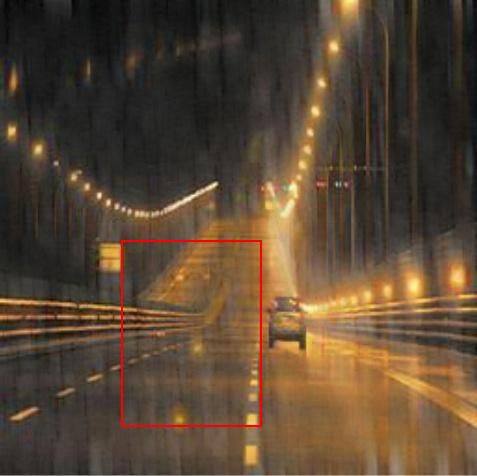}}
\end{minipage}
\hfill
\begin{minipage}{0.118\linewidth}
\centering{\includegraphics[width=1\linewidth]{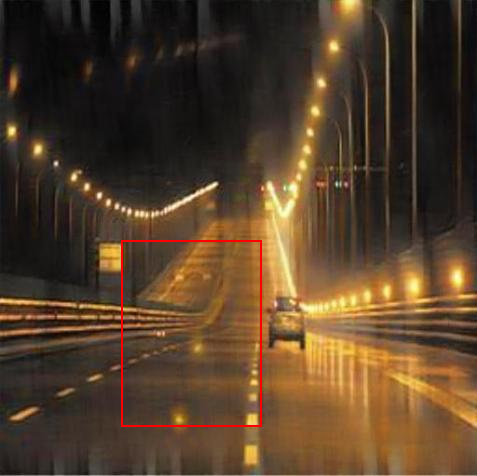}}
\end{minipage}
\hfill
\begin{minipage}{0.118\linewidth}
\centering{\includegraphics[width=1\linewidth]{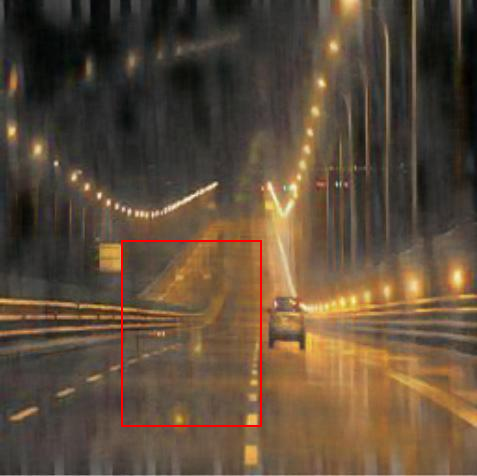}}
\end{minipage}
\hfill
\begin{minipage}{0.118\linewidth}
\centering{\includegraphics[width=1\linewidth]{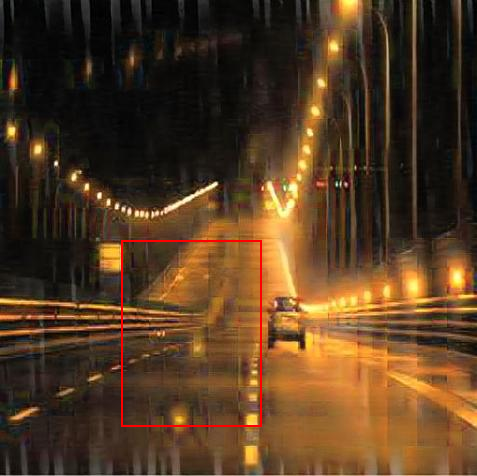}}
\end{minipage}
\hfill
\begin{minipage}{0.118\linewidth}
\centering{\includegraphics[width=1\linewidth]{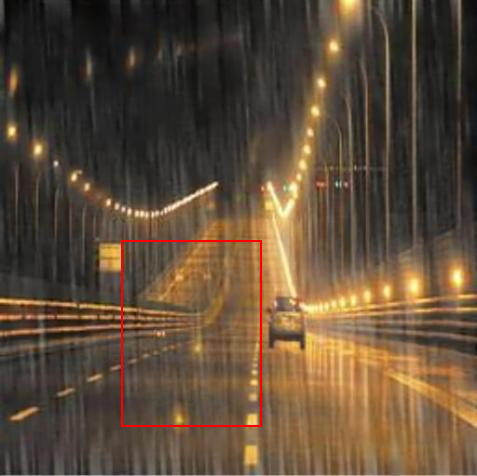}}
\end{minipage}
\hfill
\begin{minipage}{0.118\linewidth}
\centering{\includegraphics[width=1\linewidth]{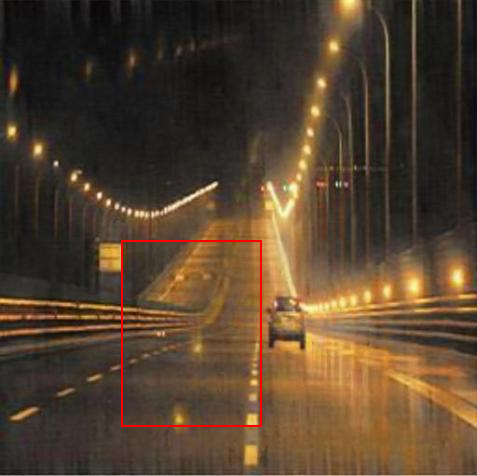}}
\end{minipage}
\vfill
\begin{minipage}{0.118\linewidth}
\centering{\includegraphics[width=1\linewidth]{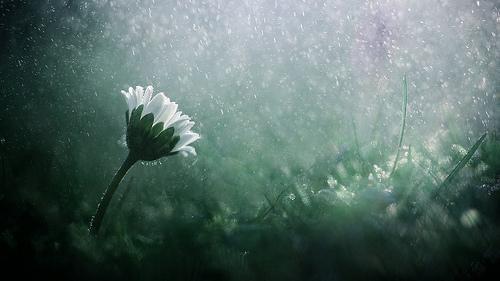}}
\centerline{(a)}
\end{minipage}
\hfill
\begin{minipage}{0.118\linewidth}
\centering{\includegraphics[width=1\linewidth]{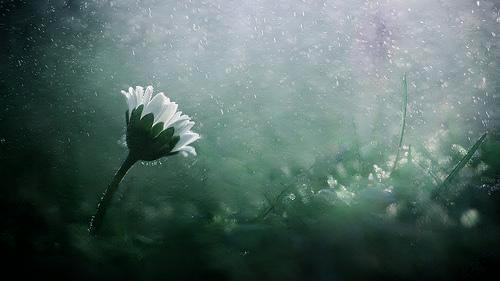}}
\centerline{(b)}
\end{minipage}
\hfill
\begin{minipage}{0.118\linewidth}
\centering{\includegraphics[width=1\linewidth]{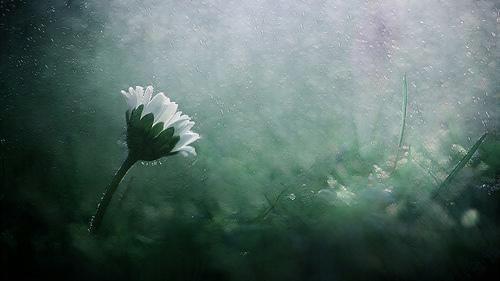}}
\centerline{(c)}
\end{minipage}
\hfill
\begin{minipage}{0.118\linewidth}
\centering{\includegraphics[width=1\linewidth]{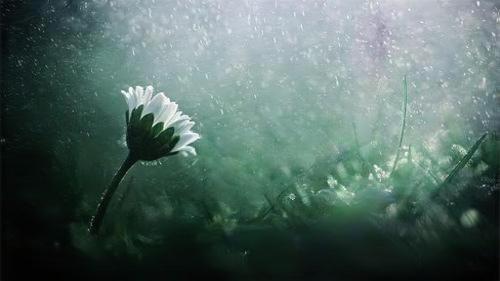}}
\centerline{(d)}
\end{minipage}
\hfill
\begin{minipage}{0.118\linewidth}
\centering{\includegraphics[width=1\linewidth]{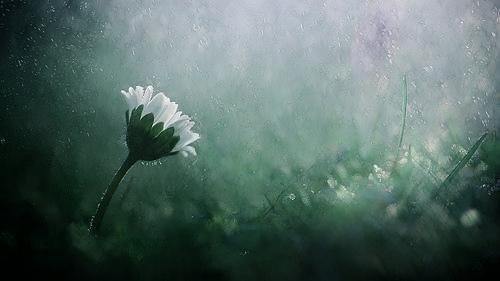}}
\centerline{(e)}
\end{minipage}
\hfill
\begin{minipage}{0.118\linewidth}
\centering{\includegraphics[width=1\linewidth]{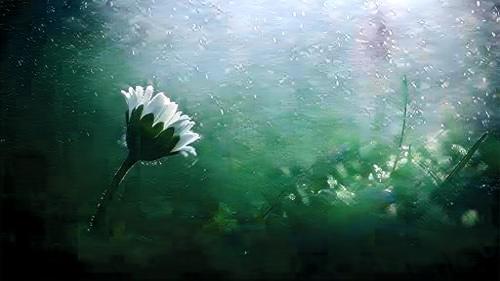}}
\centerline{(f)}
\end{minipage}
\hfill
\begin{minipage}{0.118\linewidth}
\centering{\includegraphics[width=1\linewidth]{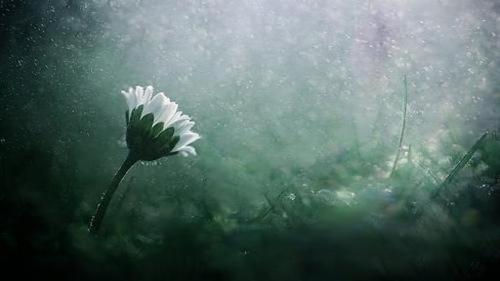}}
\centerline{(g)}
\end{minipage}
\hfill
\begin{minipage}{0.118\linewidth}
\centering{\includegraphics[width=1\linewidth]{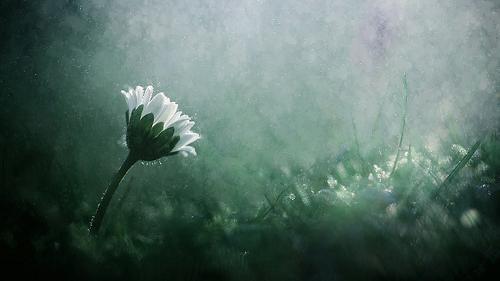}}
\centerline{(h)}
\end{minipage}
\end{center}
\caption{{Qualitative comparisons of selected methods and our method on real-world rainy images. (a) Input. (b)-(h) Deraining results of \cite{Fu_2017_CVPR}, \cite{Li_2018_ECCV}, \cite{Zhang_2018_CVPR}, \cite{Yang_wh_2019_TPAMI}, \cite{Li_rt_2019_CVPR}, \cite{Wang_ty_2019_CVPR} and our method.}}
\label{fig:practical_comparison}
\end{figure*}

\subsection{Training Loss}

We give the details of how our QSNet is trained via a logarithmic MLE based on our quasi-sparsity priors. A detail loss based on the low-value properties of rain layer $\mathbf{R}$, another $L_1$ content loss based on similarity metric and multi-scale auxiliary loss corresponding to auxiliary decoder are introduced to restore image contents.

\noindent \textbf{Quasi-sparsity loss} We assume that the rain layer $\mathbf{R}$ and background layer $\mathbf{B}$ are independent to simplify our algorithm. The quasi-sparsity priors Eq. \eqref{eq:laplacian_approximation} of rainy image $\mathbf{I}$ can be rewritten as:
\begin{equation} \label{eq:prior_define}
P_{q}(\mathbf{I})=P_{q}(\mathbf{R})P_{q}(\mathbf{B})=\prod_{i, k} \frac{\pi}{s} e^{- ( \vert \omega_{i, k} \ast \mathbf{R} \vert +  \vert \omega_{i, k} \ast \mathbf{B} \vert)/s}.
\end{equation}
By applying logarithm, Eq. \eqref{eq:prior_define} becomes:
\begin{equation} \label{eq:prior_define_log}
log(P_{q}(\mathbf{I}))=-\frac{1}{s}  \sum_{i, k} (\vert \omega_{i, k} \ast \mathbf{R} \vert + \vert \omega_{i, k} \ast \mathbf{B} \vert) + \beta.
\end{equation}
$\beta$ is a constant produced during calculating logarithm. $s$ is also constant. Hence, maximizing Eq. \eqref{eq:prior_define_log} is equal to minimizing the following loss function:
\begin{equation}\label{eq:quasi_loss_function}
\mathcal{L}_{q}=\sum_{i, k} \vert \omega_{i, k} \ast \mathbf{R} \vert + \vert \omega_{i, k} \ast \mathbf{B} \vert.
\end{equation}
For all rainy images $\{ \mathbf{I}_{t} \}^{N}_{t=1}$ in our training dataset $\{ (\mathbf{I}_{t}, \mathbf{B}_{t}) \}^{N}_{t=1}$, $\mathcal{L}_{q}$ can be rewritten as:
\begin{equation}\label{eq:quasi_loss_function_dataset}
\mathcal{L}_{Q}=\sum_{t=1}^{N}\sum_{i, k} \vert \omega_{i, k} \ast \mathcal{S}(\mathbf{I}_t) \vert + \vert \omega_{i, k} \ast [\mathbf{I}_t-\mathcal{S}(\mathbf{I}_t)] \vert.
\end{equation}

Clearly, quasi-sparsity is formed by adding several first or second order derivatives, which contains more complete image texture information, an example is in Fig. \ref{fig:sparse_space_image}. However, rain streaks cannot be decomposed only via constraint on textures, \ie deraining task is still ill-posed. 

\noindent \textbf{Content loss} We utilize MAE to calculate the difference between deraining result and ground truth background to recover image contents. It can be written as:
\begin{equation}\label{eq:content_loss_function}
\mathcal{L}_{C}=\sum_{t=1}^{N} | \mathbf{I}_{t}-\mathcal{S}(\mathbf{I}_{t})-\mathbf{B}_{t} |.
\end{equation}

\noindent \textbf{Detail loss} As shown in Fig. \ref{fig:sparse_space_image}(c), the non-rain areas have very low values in $\mathbf{R}$, which is formulated by a detail loss to restore image details:
\begin{equation}\label{eq:detail_loss_function}
\mathcal{L}_{D}= | (\mathds{1}-\mathbf{L}) \circ \mathbf{R} |,
\end{equation}
where $\mathbf{L}$ is the location map of rain streaks obtain by \cite{Yang_2017_CVPR}. $\mathds{1}$ is all-one matrix.

\noindent \textbf{Multi-scale auxiliary loss} The role of different auxiliary decoder is to keep content similarity. MSE is used as basic metric. Assume $\{\mathcal{A}_{i}(\cdot)\}^{5}_{i=1}$ denote the five auxiliary decoders, our multi-scale auxiliary loss is defined as follows:
\begin{equation}\label{eq:auxiliary_loss_function}
\mathcal{L}_{A}=\frac{1}{5}\sum^{5}_{i=1}\sum_{t=1}^{N} \| \mathbf{I}_{t}-\mathcal{A}_{i}(\mathbf{F}_{t,i})-\mathbf{B}_{t} \|^{2}_{F},
\end{equation}
where $\mathbf{F}_{t,i}$ is the feature map of $\mathbf{I}_{t}$ input into $\mathcal{A}_i(\cdot)$.
Our whole loss function is:
\begin{equation}\label{eq:whole_loss_function}
\mathcal{L}= \lambda_{Q} \mathcal{L}_{Q} + \lambda_{C} \mathcal{L}_{C} + \lambda_{A} \mathcal{L}_{A} + \lambda_{D} \mathcal{L}_{D}
\end{equation}

\begin{table*}[]
\centering
\caption{Average running time comparisons of selected methods and our method. The image size is $512 \times 512$.}
\begin{tabular}{c|cccccc|c|c}
\hline
\hline
 Methods & \cite{Fu_2017_CVPR} & \cite{Li_2018_ECCV} & \cite{Zhang_2018_CVPR} & \cite{Yang_wh_2019_TPAMI} & \cite{Li_rt_2019_CVPR} & \cite{Wang_ty_2019_CVPR} & w/o Sharing & Ours \\ \hline
 \hline
 Time & $0.09s$ & $0.47s$ & $0.06s$ & $1.39s$ & $0.45s$ & $0.66s$ & $0.007s$ & $0.005s$ \\ \hline
 \hline
\end{tabular}
\label{tab:time_comparison}
\end{table*}

\begin{table*}[]
\centering
\caption{PSNR/SSIM of deraining results decoded from different scale features.}
\label{tab:multi-scale}
\begin{tabular}{c|c|c|c|c|c|c}
\hline
\hline
 Scales & $C_{1}$ & $C_{2}$ & $C_{3}$ & $C_{4}$ & $C_{5}$  & All scales \\ \hline
 \hline
 Test-I & 31.68/0.902 & 32.14/0.907 & 32.18/0.902 & 30.75/0.862 & 30.69/0.870 & \textbf{33.15/0.923} \\ 
 Test-II & 23.86/0.738 & 24.85/0.749 & 24.91/0.774 & 23.86/0.687 & 22.66/0.669 & \textbf{25.66/0.830} \\
 Test-III & 31.80/0.909 & 32.56/0.911 & 32.67/0.920 & 31.48/0.871 & 31.16/0.877 & \textbf{33.94/0.938} \\ \hline
 \hline
\end{tabular}
\end{table*}


\section{Experiments}
\label{sec:ExperimentalResults}

In order to evaluate the performance of our method, PSNR and SSIM \cite{Wang_2004_TIP} are selected as objective metrics. Three state-of-the-art works \cite{Zhang_2018_CVPR,Li_2018_ECCV,Fu_2017_CVPR} and three very recent works \cite{Yang_wh_2019_TPAMI,Li_rt_2019_CVPR,Wang_ty_2019_CVPR} are selected to make comparisons with our method. 

\subsection{Implementation Details}

The training $256 \times 256$ patch pairs are randomly cropped from our training dataset. Adam \cite{Kingma_2015_ICLR} is utilized as our optimizer. The learning rate is set to $0.001$ initially and decreases by multiplying $0.1$ when the loss stops improving. Our code is implemented on a NVIDIA 1080Ti GPU based on Pytorch. The parameters $\lambda_{Q}$, $\lambda_{C}$, $\lambda_{A}$ and $\lambda_{D}$ in Eq. \eqref{eq:whole_loss_function} are $10^{-3}$, $1$, $0.01$ and $10^{-4}$ respectively.

\subsection{Dataset}

We use the training dataset in \cite{Li_2018_arxiv} as our training dataset. We randomly select $100$ pairs from the testing datasets of \cite{Zhang_2018_CVPR,Zhang_2017_arxiv_derain_gan,Fu_2017_CVPR} respectively to constitute our first testing dataset which covers commonly used benchmark testing datasets (Test-I). We utilize the Rain100H by Yang \textit{et al.} \cite{Yang_2017_CVPR} as our second testing dataset (Test-II). Our third testing dataset (Test-III) consists of rainy images with wide streaks and blur edges, including $400$ testing pairs.

\begin{figure}[t!]
\begin{minipage}{0.132\linewidth}
\centering{\includegraphics[width=1\linewidth]{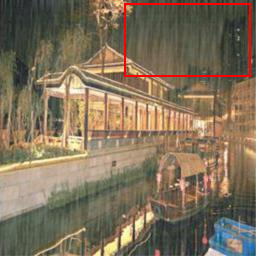}}
\end{minipage}
\hfill
\begin{minipage}{0.132\linewidth}
\centering{\includegraphics[width=1\linewidth]{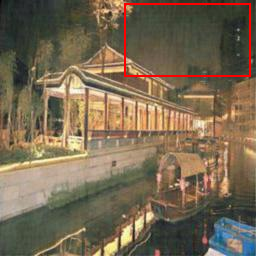}}
\end{minipage}
\hfill
\begin{minipage}{0.132\linewidth}
\centering{\includegraphics[width=1\linewidth]{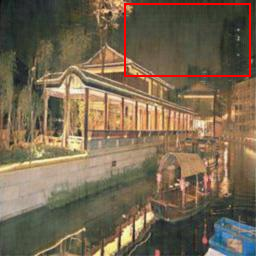}}
\end{minipage}
\hfill
\begin{minipage}{0.132\linewidth}
\centering{\includegraphics[width=1\linewidth]{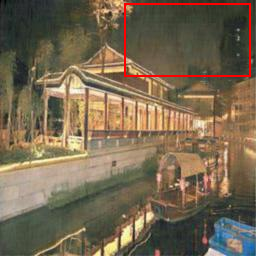}}
\end{minipage}
\hfill
\begin{minipage}{0.132\linewidth}
\centering{\includegraphics[width=1\linewidth]{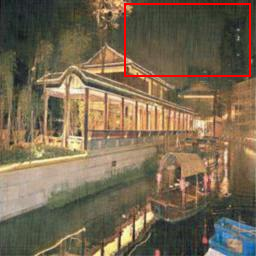}}
\end{minipage}
\hfill
\begin{minipage}{0.132\linewidth}
\centering{\includegraphics[width=1\linewidth]{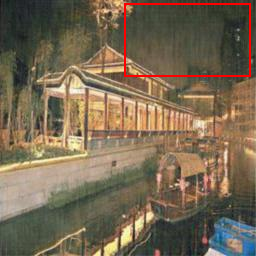}}
\end{minipage}
\hfill
\begin{minipage}{0.132\linewidth}
\centering{\includegraphics[width=1\linewidth]{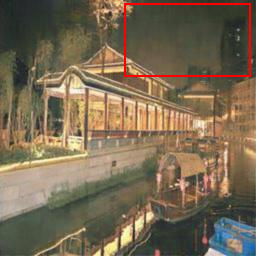}}
\end{minipage}
\vfill
\begin{minipage}{0.132\linewidth}
\includegraphics[width=1\linewidth]{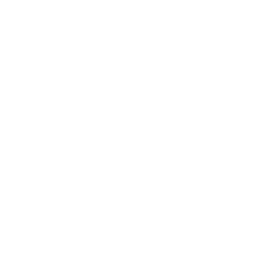}
\centerline{(a)}
\end{minipage}
\hfill
\begin{minipage}{0.132\linewidth}
\includegraphics[width=1\linewidth]{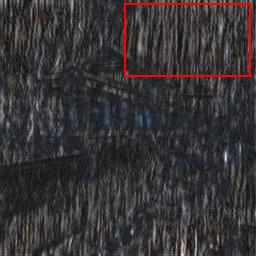}
\centerline{(b)}
\end{minipage}
\hfill
\begin{minipage}{0.132\linewidth}
\includegraphics[width=1\linewidth]{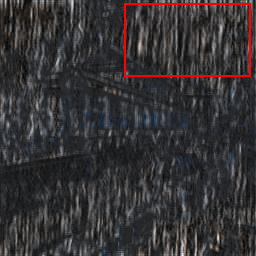}
\centerline{(c)}
\end{minipage}
\hfill
\begin{minipage}{0.132\linewidth}
\includegraphics[width=1\linewidth]{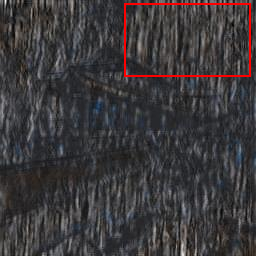}
\centerline{(d)}
\end{minipage}
\hfill
\begin{minipage}{0.132\linewidth}
\includegraphics[width=1\linewidth]{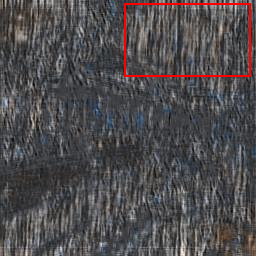}
\centerline{(e)}
\end{minipage}
\hfill
\begin{minipage}{0.132\linewidth}
\includegraphics[width=1\linewidth]{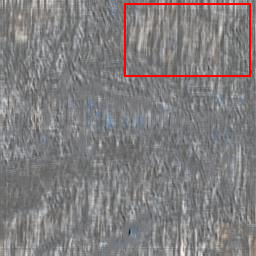}
\centerline{(f)}
\end{minipage}
\hfill
\begin{minipage}{0.132\linewidth}
\includegraphics[width=1\linewidth]{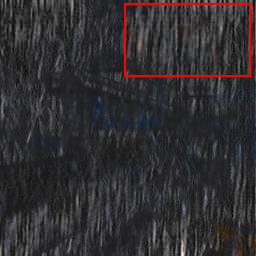}
\centerline{(g)}
\end{minipage}
\caption{Deraining results decoded from different scale features. (a) Input. (b)-(g) Deraining results decoded from $C_{1}$, $C_{2}$, $C_{3}$, $C_{4}$, $C_{5}$ and all scales.}
\label{fig:multi-scale}
\end{figure}

\subsection{Quantitative Evaluation on Synthetic Datasets}

We show the PSNR/SSIM values of different methods in Table \ref{tab:psnr_ssim_comparison}. Our method outperforms other state-of-the-art methods. On Test-I and Test-III, our method obtains 3 dB higher PSNR values than the second best method. In Fig. \ref{fig:synthetic_compare}, we visually show two synthetic rainy images. The deraining results of our method are most similar to the ground truth. The method \cite{Zhang_2018_CVPR} also produces good result for the first rainy image, but this method makes the cat face more white than the ground truth. For the second heavy rain image our method remove rain better. The objective indexes of \cite{Li_rt_2019_CVPR} are lowest, as it always changes the color hue of deraining results.

\subsection{Qualitative Evaluation on Real-World Images}

Visual comparisons with the state-of-the-art methods on real-world rainy images are shown in Fig. \ref{fig:practical_comparison}. For light rainy image (e.g., the first one) all the method can obtain good results, except that some works \cite{Li_2018_ECCV,Yang_wh_2019_TPAMI,Li_rt_2019_CVPR} will lose image details. For heavy rainy image (\eg, the second one) our method obtains best results. The work \cite{Li_rt_2019_CVPR} removes majority of rain streaks, but this work will produce blocking effect for some rainy images. Another problem of this work is that its results sometimes are dark, that important image details will be lost (\eg, the third one) and the colors of original rainy images will change. The works \cite{Li_2018_ECCV,Yang_wh_2019_TPAMI,Wang_ty_2019_CVPR} obtain very good results for the third rainy images, but our results still outperforms theirs.

\subsection{Running Time}

We show the running time of different methods in Table \ref{tab:time_comparison}. All the methods are tested on the same GPU. We can see that our method possesses the fastest running speed and is one order of magnitude faster than the second fastest method \cite{Zhang_2018_CVPR}. The average running speed of our method is $0.005s$ per $512 \times 512$ image, \ie, our method can handle $200$ $512 \times 512$ images per second. This is attributed to the group convolution and feature exchanging among different groups and different scale spaces, which accelerates the propagation of information in the network \cite{Zhang_2018_CVPR_Shuffle}.

\begin{figure}[t!]
\begin{minipage}{0.155\linewidth}
\centering{\includegraphics[width=1\linewidth]{images/rain-20.jpg}}
\end{minipage}
\hfill
\begin{minipage}{0.155\linewidth}
\centering{\includegraphics[width=1\linewidth]{images/rain-20_sparsity.jpg}}
\end{minipage}
\hfill
\begin{minipage}{0.155\linewidth}
\centering{\includegraphics[width=1\linewidth]{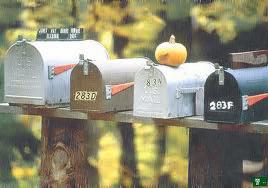}}
\end{minipage}
\hfill
\begin{minipage}{0.155\linewidth}
\centering{\includegraphics[width=1\linewidth]{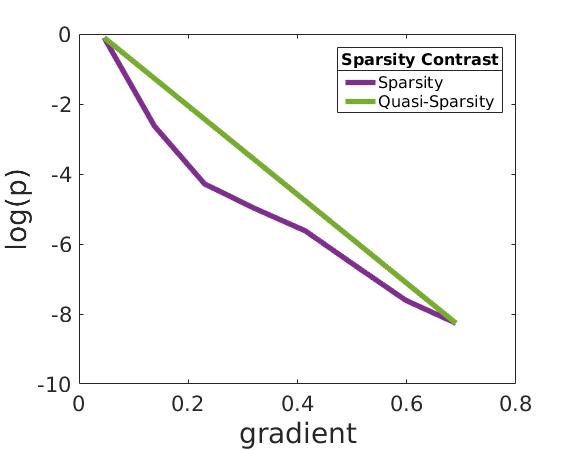}}
\end{minipage}
\hfill
\begin{minipage}{0.155\linewidth}
\centering{\includegraphics[width=1\linewidth]{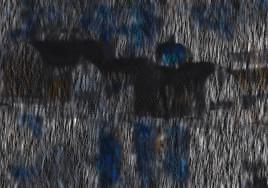}}
\end{minipage}
\hfill
\begin{minipage}{0.155\linewidth}
\includegraphics[width=1\linewidth]{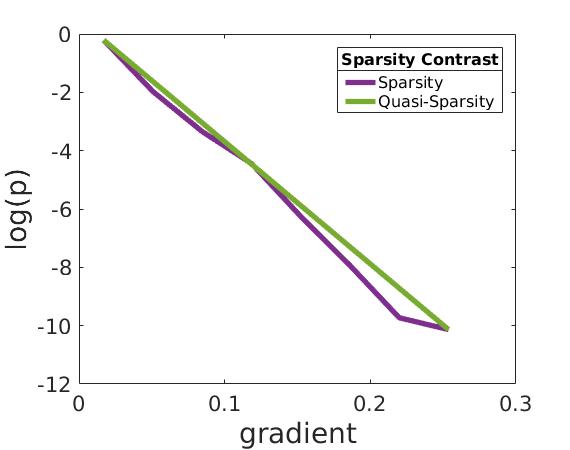}
\end{minipage}
\vfill
\begin{minipage}{0.155\linewidth}
\centering{\includegraphics[width=1\linewidth]{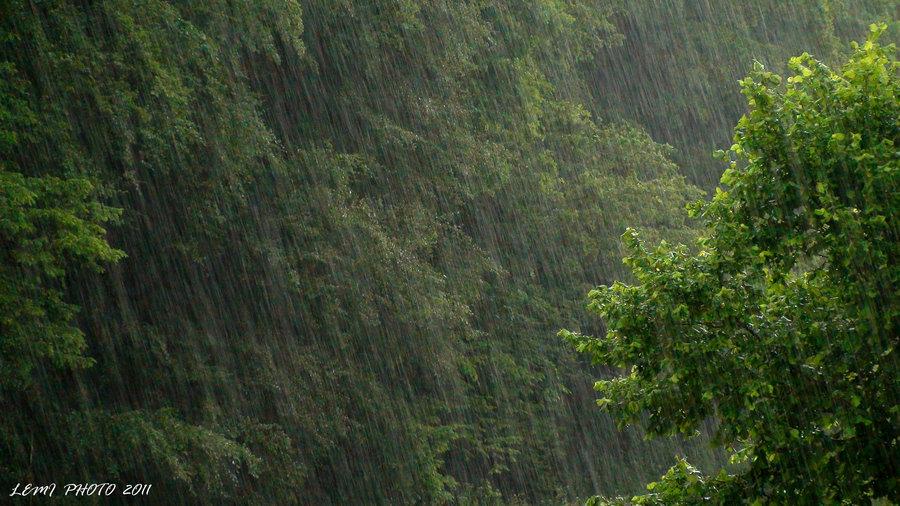}}
\centerline{(a)}
\end{minipage}
\hfill
\begin{minipage}{0.155\linewidth}
\centering{\includegraphics[width=1\linewidth]{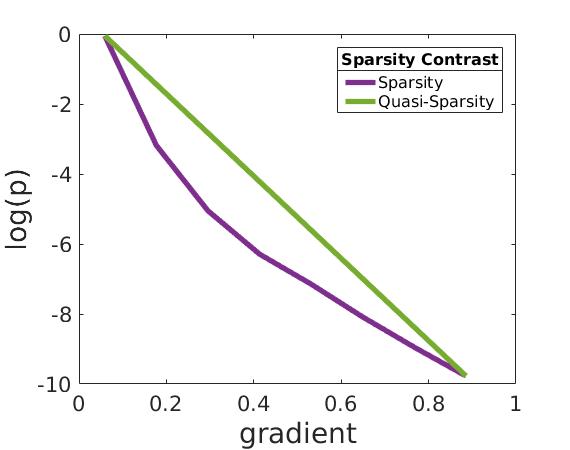}}
\centerline{(b)}
\end{minipage}
\hfill
\begin{minipage}{0.155\linewidth}
\centering{\includegraphics[width=1\linewidth]{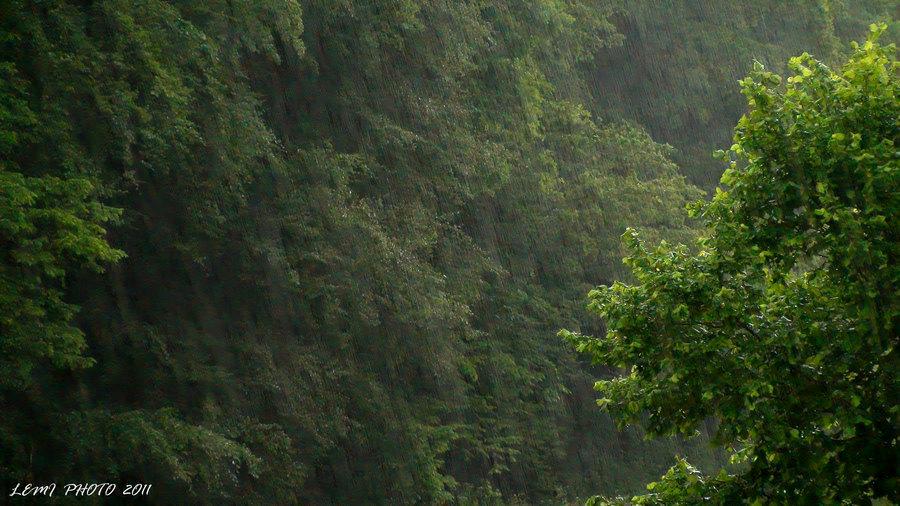}}
\centerline{(c)}
\end{minipage}
\hfill
\begin{minipage}{0.155\linewidth}
\centering{\includegraphics[width=1\linewidth]{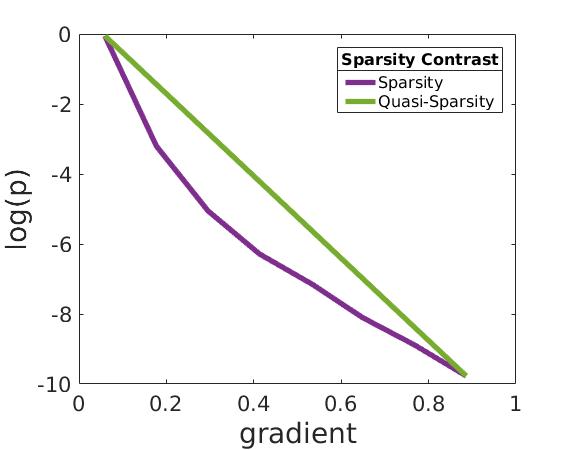}}
\centerline{(d)}
\end{minipage}
\hfill
\begin{minipage}{0.155\linewidth}
\centering{\includegraphics[width=1\linewidth]{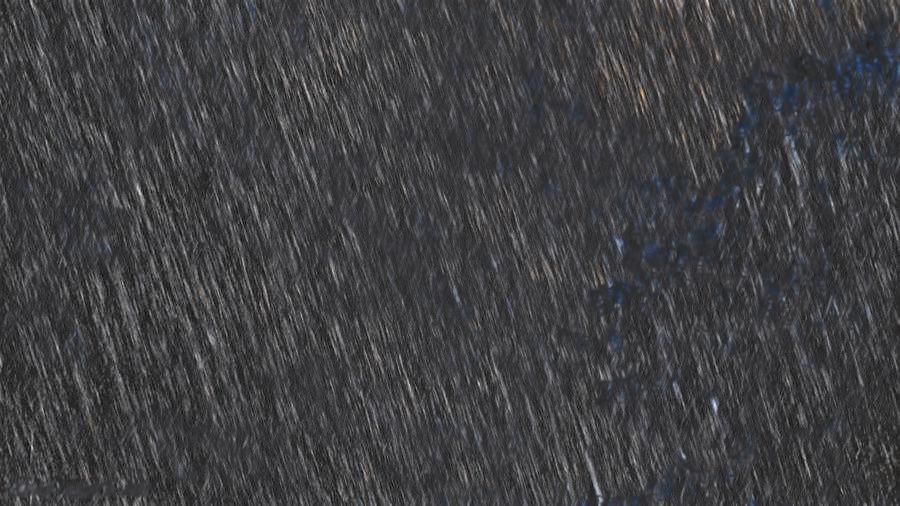}}
\centerline{(e)}
\end{minipage}
\hfill
\begin{minipage}{0.155\linewidth}
\includegraphics[width=1\linewidth]{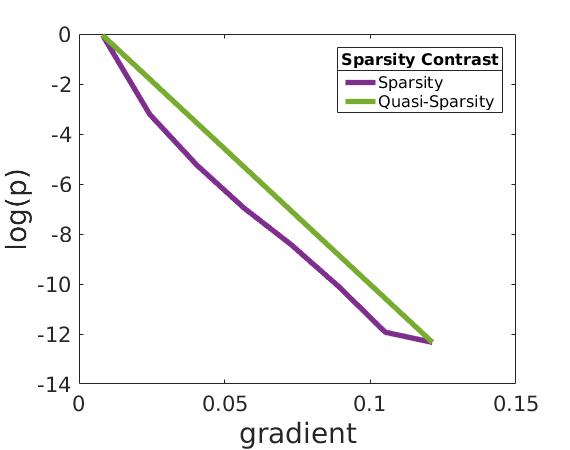}
\centerline{(f)}
\end{minipage}
\caption{Sparsity of real-world rainy images, the deraining results and the decomposed rainy layers.}
\label{fig:sparsity}
\end{figure}

\subsection{Multi-scale Decoding}

In this subsection, we study the features of different scales via qualitative and quantitative evaluation to illustrate which kinds of features contribute most to image deraining. In the multi-scale feature extraction part, $C_{1}$ denotes the convolution layer with $1 \times 1$ kernel, $C_{2}$, $C_{3}$, $C_{4}$, $C_{5}$ denote the atrous convolution with atrous rate equalling to $1$, $2$, $4$, $6$ respectively. In Fig. \ref{fig:multi-scale}, we visually show the results of multi-scale decoding of a real-world rainy image. We can see that some traces of rain streaks with large size appear in the results decoded from small-scale features, \eg, Fig. \ref{fig:multi-scale}(b). Similarly, some slim rain streaks also appear in the results decoded from large-scale features \eg, Fig. \ref{fig:multi-scale}(e). The result decoded by multi-scale features possesses best visual quality and clearest background. This is because small-scale convolution captures features of small rain streaks and tends to neglect features of some large rain streaks, vice versa. When combined, different-scale convolutions remove corresponding-scale rain streaks.

To analyze the role of different-scale features more accurately, we show the PSNR/SSIM values of different-scale decoding results on our three testing datasets in Table \ref{tab:multi-scale}. $C_{2}$ and $C_{3}$ (atrous rates are $1$ and $2$) produce the higher PSNR/SSIM values, illustrating that majority of rain streaks possess the width of $3$ or $5$ pixels in common images. Of course, the best results are obtained by fusing multi-scale features together.

\subsection{Sparsity}
We utilize a logarithmic maximum likelihood estimation based on the sparsity priors of rainy images to train our network. To facilitate this process, we formulate image sparsity as quasi-sparsity priors. Quasi-sparsity is after all not sparsity, so that other three cost functions $\mathcal{L}_{C}$, $\mathcal{L}_{A}$ and $\mathcal{L}_{D}$ are introduced to force quasi-sparsity to sparsity. We verify the sparsity of deraining results and rainy layers learned by our QSNet. In Fig. \ref{fig:sparsity}, we show the sparsity of two real-world rainy images, their deraining resuls and rain layers. We can see that obtained experimental results are consistent with our previous theoretical analysis.

\subsection{Ablation Study}

\begin{figure}[t!]
\begin{minipage}{0.19\linewidth}
\centering{\includegraphics[width=1\linewidth]{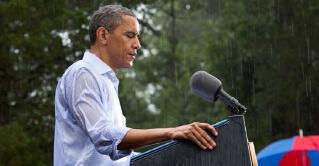}}
\end{minipage}
\hfill
\begin{minipage}{0.19\linewidth}
\centering{\includegraphics[width=1\linewidth]{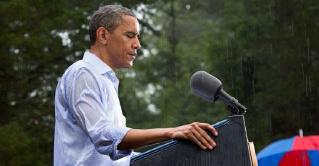}}
\end{minipage}
\hfill
\begin{minipage}{0.19\linewidth}
\centering{\includegraphics[width=1\linewidth]{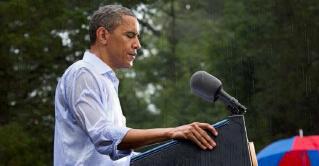}}
\end{minipage}
\hfill
\begin{minipage}{0.19\linewidth}
\centering{\includegraphics[width=1\linewidth]{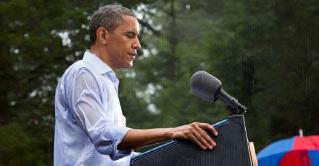}}
\end{minipage}
\hfill
\begin{minipage}{0.19\linewidth}
\centering{\includegraphics[width=1\linewidth]{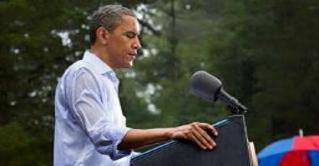}}
\end{minipage}
\vfill
\begin{minipage}{0.19\linewidth}
\includegraphics[width=1\linewidth]{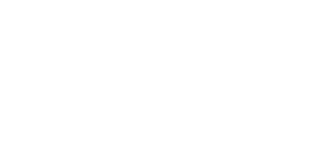}
\centerline{(a)}
\end{minipage}
\hfill
\begin{minipage}{0.19\linewidth}
\includegraphics[width=1\linewidth]{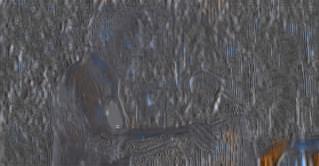}
\centerline{(b)}
\end{minipage}
\hfill
\begin{minipage}{0.19\linewidth}
\includegraphics[width=1\linewidth]{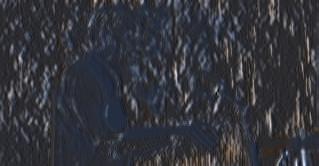}
\centerline{(c)}
\end{minipage}
\hfill
\begin{minipage}{0.19\linewidth}
\includegraphics[width=1\linewidth]{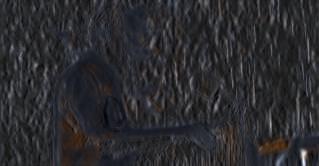}
\centerline{(d)}
\end{minipage}
\hfill
\begin{minipage}{0.19\linewidth}
\includegraphics[width=1\linewidth]{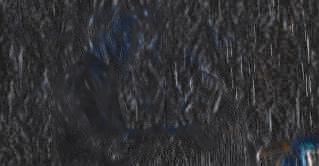}
\centerline{(e)}
\end{minipage}
\caption{Visual results of ablation studies. (a) Input. (b)-(e) Results of $V_{1}$, $V_{2}$, $V_{3}$ and $V_{4}$. The second line is the rain layers.}
\label{fig:ablation}
\end{figure}

Our network is optimized by an end-to-end training process and four robust loss functions are developed to enhance network performance by deeply exploring intrinsic image properties. In order to study the contributions of different losses to network performance, we test network by ablating the four loss functions one-by-one. We utilize four different methods to train our QSNet: 1) only content loss $\mathcal{L}_{C}$ is used to update network parameters and named as $V_{1}$; 2) $\mathcal{L}_{C} + \mathcal{L}_{Q}$ is used and named as $V_{2}$; 3) $\mathcal{L}_{C} + \mathcal{L}_{Q} + \mathcal{L}_{D}$ is used and named as $V_{3}$; 4) $\mathcal{L}_{C} + \mathcal{L}_{Q} + \mathcal{L}_{D} + \mathcal{L}_{A}$ is used and named as $V_{4}$. In Table \ref{tab:ablation}, we show the PSNR and SSIM evaluation on our three testing datasets. Each loss promotes our network performance substantially. In Fig. \ref{fig:ablation}, we utilize a real-world rainy image to visually show the results of our ablation studies. We can see that rain streaks become less and less after the loss functions is introduced one-by-one. In the rainy layer, the image details are reduced step-wisely and the rain streaks increase accordingly.  

\begin{table}[]
\centering
\caption{PSNR/SSIM of the different variants of our method.}
\label{tab:ablation}
\begin{tabular}{c|c|c|c|c}
\hline
\hline
     Variants             & $V_{1}$ & $V_{2}$ & $V_{3}$ & $V_{4}$ \\ \hline
     \hline
     $\mathcal{L}_{C}$ ?  & $\surd$ & $\surd$ & $\surd$ & $\surd$ \\
     $\mathcal{L}_{Q}$ ?  & w/o & $\surd$ & $\surd$ & $\surd$ \\
     $\mathcal{L}_{D}$ ?  & w/o & w/o & $\surd$ & $\surd$ \\
     $\mathcal{L}_{A}$ ?  & w/o & w/o & w/o & $\surd$ \\ \hline
     \hline
\multirow{2}{*}{Test-I} & 29.94 & 31.70 & 32.03 & \textbf{33.15} \\
                  & 0.858 & 0.898 & 0.908 & \textbf{0.923} \\ \hline
\multirow{2}{*}{Test-II} & 24.75 & 25.21 & 25.55 & \textbf{25.66} \\
                  & 0.716 & 0.773 & 0.791 & \textbf{0.830} \\ \hline
\multirow{2}{*}{Test-III} & 31.19 & 32.61 & 32.91 & \textbf{33.94} \\
                  & 0.883 & 0.913 & 0.921 & \textbf{0.938} \\ \hline
                  \hline
\end{tabular}
\end{table}

\subsection{Studies of Information Sharing}

\begin{figure}[t!]
\begin{minipage}{0.155\linewidth}
\centering{\includegraphics[width=1\linewidth]{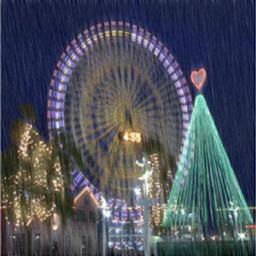}}
\end{minipage}
\hfill
\begin{minipage}{0.155\linewidth}
\centering{\includegraphics[width=1\linewidth]{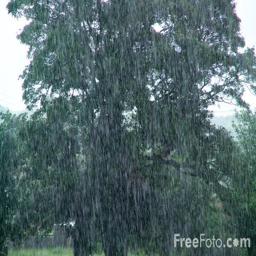}}
\end{minipage}
\hfill
\begin{minipage}{0.155\linewidth}
\centering{\includegraphics[width=1\linewidth]{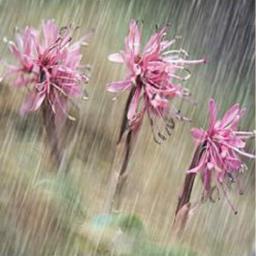}}
\end{minipage}
\hfill
\begin{minipage}{0.155\linewidth}
\centering{\includegraphics[width=1\linewidth]{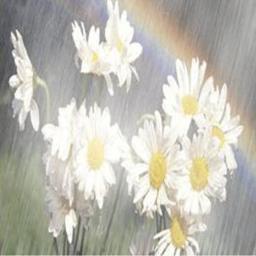}}
\end{minipage}
\hfill
\begin{minipage}{0.155\linewidth}
\centering{\includegraphics[width=1\linewidth]{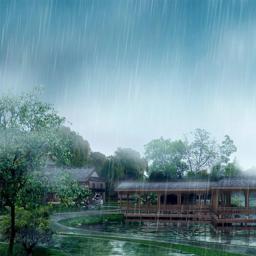}}
\end{minipage}
\hfill
\begin{minipage}{0.155\linewidth}
\includegraphics[width=1\linewidth]{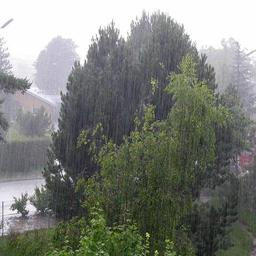}
\end{minipage}
\vfill
\begin{minipage}{0.155\linewidth}
\centering{\includegraphics[width=1\linewidth]{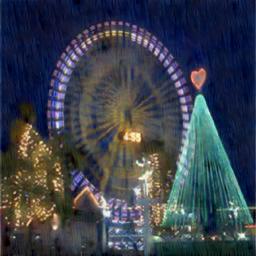}}
\end{minipage}
\hfill
\begin{minipage}{0.155\linewidth}
\centering{\includegraphics[width=1\linewidth]{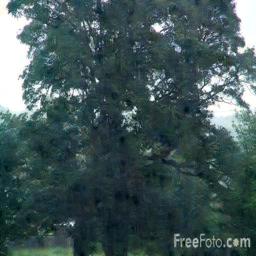}}
\end{minipage}
\hfill
\begin{minipage}{0.155\linewidth}
\centering{\includegraphics[width=1\linewidth]{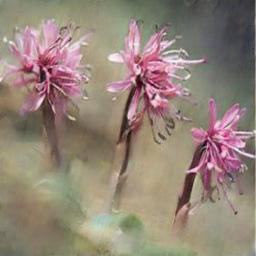}}
\end{minipage}
\hfill
\begin{minipage}{0.155\linewidth}
\centering{\includegraphics[width=1\linewidth]{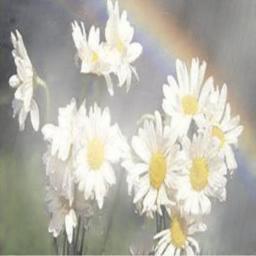}}
\end{minipage}
\hfill
\begin{minipage}{0.155\linewidth}
\centering{\includegraphics[width=1\linewidth]{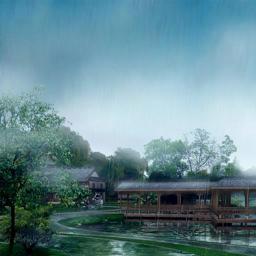}}
\end{minipage}
\hfill
\begin{minipage}{0.155\linewidth}
\includegraphics[width=1\linewidth]{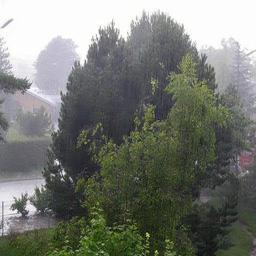}
\end{minipage}
\vfill
\begin{minipage}{0.155\linewidth}
\centering{\includegraphics[width=1\linewidth]{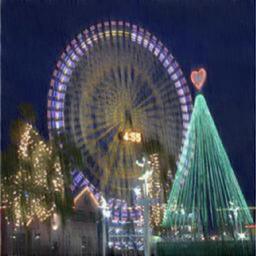}}
\centerline{(a)}
\end{minipage}
\hfill
\begin{minipage}{0.155\linewidth}
\centering{\includegraphics[width=1\linewidth]{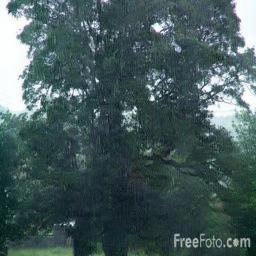}}
\centerline{(b)}
\end{minipage}
\hfill
\begin{minipage}{0.155\linewidth}
\centering{\includegraphics[width=1\linewidth]{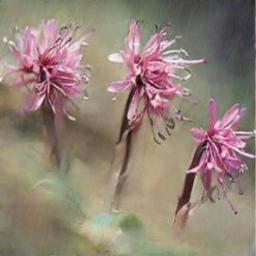}}
\centerline{(c)}
\end{minipage}
\hfill
\begin{minipage}{0.155\linewidth}
\centering{\includegraphics[width=1\linewidth]{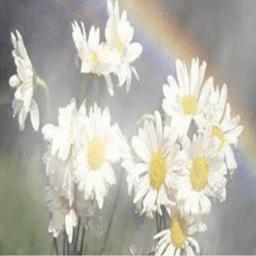}}
\centerline{(d)}
\end{minipage}
\hfill
\begin{minipage}{0.155\linewidth}
\centering{\includegraphics[width=1\linewidth]{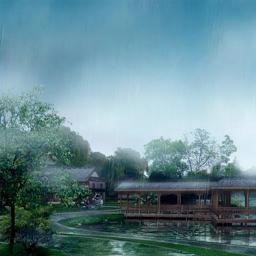}}
\centerline{(e)}
\end{minipage}
\hfill
\begin{minipage}{0.155\linewidth}
\includegraphics[width=1\linewidth]{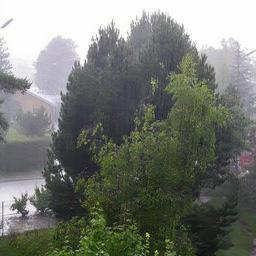}
\centerline{(f)}
\end{minipage}
\caption{The first line is input rainy images. The second and third lines are deraining results w/o or w/ information sharing.}
\label{fig:no_sharing}
\end{figure}

Our QSNet exchanges features from different convolution groups or different scales. In Table \ref{tab:sharing}, we show the objective indexes with and without information sharing. We can see that the PSNR/SSIM values on our three testing datasets all decrease, which proves the importance of feature sharing in deraining neural networks. Table \ref{tab:time_comparison} also shows that the running speed decreases by $0.002s$ without feature sharing. Some visual results are shown in Fig. \ref{fig:no_sharing}. The first two real-world rainy images are two failure cases without information sharing, where some black blocks appear. This is may be due to the high outlier values produced in $\mathbf{R}$ during feature fusing without information sharing.

\begin{table}[]
\centering
\caption{PSNR/SSIM of our QSNet w/o or w/ information sharing.}
\label{tab:sharing}
\begin{tabular}{c|c|c|c}
\hline
\hline
Datasets & Test-I & Test-II & Test-III \\ \hline
\hline
w/o Sharing & 31.91/0.897 & 25.53/0.799 & 32.64/0.907 \\ \hline
QSSD & 33.15/0.923 & 25.66/0.830 & 33.94/0.938 \\ \hline
\hline
\end{tabular}
\end{table}

\section{Conclusions}
\label{sec:Conclusion}
Sparsity priors are intrinsic properties of unbroken natural images. They reflect the integrality of image textures by making statistics for the log-histogram of images in different gradient domains, which is favorable to unbroken image decomposition, as well as image deraining. In this paper, we determine the sparsity priors of rainy images via a robust statistic. To let our deraining results robustly possess such properties and restore intact textures, quasi-sparsity priors are developed to ease the network training via maximum likelihood estimation based on image sparsity. Moreover, a similarity metric loss and low-value loss are followed to restore image contents. We also study the multi-scale features via an auxiliary decoding structure. Accordingly, an auxiliary optimization loss improves deraining performance further. Quantitative and qualitative evaluations both illustrate that our method outperforms the state-of-the-art.


\end{document}